\documentclass{article}

\PassOptionsToPackage{numbers,compress}{natbib}

\usepackage[sglblindworkshop,final]{neurips_2025}
\makeatletter
\renewcommand{\@notice}{}
\makeatother

\usepackage[utf8]{inputenc}
\usepackage[T1]{fontenc}
\usepackage{graphicx}
\usepackage{booktabs}
\usepackage{amsmath}
\usepackage{enumitem}
\usepackage{multirow}
\usepackage{tabularx}
\usepackage{float}
\usepackage{subcaption}
\usepackage{xcolor}
\usepackage{url}
\usepackage{microtype}
\usepackage{listings}
\definecolor{codebg}{rgb}{0.95,0.95,0.95}
\definecolor{codegreen}{rgb}{0.0,0.4,0.0}
\definecolor{codegray}{rgb}{0.5,0.5,0.5}
\definecolor{codepurple}{rgb}{0.4,0.0,0.6}
\lstdefinestyle{pythonstyle}{
    backgroundcolor=\color{codebg},
    basicstyle=\ttfamily\scriptsize,
    keywordstyle=\color{codepurple}\bfseries,
    stringstyle=\color{codegreen},
    commentstyle=\color{codegray}\itshape,
    breaklines=true,
    frame=single,
    framerule=0.4pt,
    rulecolor=\color{codegray},
    numbers=none,
    xleftmargin=2pt,
    xrightmargin=2pt,
    aboveskip=6pt,
    belowskip=6pt
}
\lstdefinestyle{jsonstyle}{
    backgroundcolor=\color{codebg},
    basicstyle=\ttfamily\scriptsize,
    breaklines=true,
    frame=single,
    framerule=0.4pt,
    rulecolor=\color{codegray},
    numbers=none,
    xleftmargin=2pt,
    xrightmargin=2pt,
    aboveskip=6pt,
    belowskip=6pt
}
\lstdefinestyle{textstyle}{
    backgroundcolor=\color{codebg},
    basicstyle=\ttfamily\scriptsize,
    breaklines=true,
    frame=single,
    framerule=0.4pt,
    rulecolor=\color{codegray},
    numbers=none,
    xleftmargin=2pt,
    xrightmargin=2pt,
    aboveskip=6pt,
    belowskip=6pt
}

\providecommand{\textsc}[1]{\textsf{\footnotesize #1}}
\renewcommand{\textsc}[1]{{\scshape #1}}

\title{What the Author Wrote, What the Agent Reads:\\Silent Cue Loss in Enterprise Document Ingestion}

\author{%
  Yikai Zhao \\
  Amazon \\
  \texttt{yikai@amazon.com} \\
  \And
  Saurabh Pandey \\
  Amazon \\
  \texttt{saurabpd@amazon.com} \\
  \And
  Pradeep Kumar Misra \\
  Amazon \\
  \texttt{pkmisra@amazon.com} \\
}

\begin{document}

\maketitle

\begin{abstract}
Enterprise agents act on documents. Before an agent can follow a policy, answer a compliance
question or file a ticket, some ingestion layer must turn a Word file or PDF into tokens---and
that layer is where an agent deployed in the wild silently loses information. Mature converters
recover text, tables and structure faithfully, so the pipeline looks healthy. We show that one
layer is lost by all of them---the meaning an author encodes in how text \emph{looks}---and that
the loss makes agents \emph{confidently wrong} rather than visibly incomplete, which is the
harder failure to detect in deployment.

We build a task from 76 real policy documents: recover the requirements the author singled out
for attention, scored against the file's own OOXML properties by a deterministic matcher with no
LLM judge. Formatting-blind ingestion collapses from micro F1 $0.690$ to $0.150$ (docling) and
$0.075$ (plain text), a gap that replicates across \emph{six} answering models from two vendors---five Claude
models spanning the Opus, Sonnet and Haiku tiers, and OpenAI's GPT-5.6. The failure is silent in a way that matters for autonomy: an
agent that cannot see the cue guesses, and docling's false-positive rate on requirements the
author did \emph{not} flag is four times ours.

A span-level leave-one-out localises the loss to two cues---highlight ($+0.302$) and colour
($+0.262$), both $p_{\text{Holm}}<10^{-4}$---while bold, italic, monospace and strikethrough each
sit within $\pm0.003$ of zero. The same asymmetry appears in PDF. Colour is also the cue no
mainstream converter preserves and no parsing benchmark scores, and it is common: $32.8\%$ of
$6{,}460$ documents carry a deliberate colour or highlight, $10.3\%$ on an obligation. A
pre-registered test shows that \emph{typing} a cue for the agent adds $+0.085$ beyond merely
preserving it ($p_{\text{Holm}}=0.049$). We validate the labels against rendered page images
through an independent modality ($\kappa=0.72$) and bound what wording alone can achieve
(AUC $0.664$). For agent builders the practical finding is that ingestion, not reasoning, is
where this reliability is won or lost, and that the fix is cheap and measurable.
\end{abstract}

\section{Introduction}
\label{sec:intro}

An enterprise agent deployed in the wild does not read documents; it reads whatever an ingestion
layer hands it. Before it can answer a compliance question, follow an operating procedure, or decide
whether an action is permitted, some converter has turned a \texttt{.docx} or PDF into tokens. That
layer is rarely the object of study, and it is where a large amount of an agent's practical
reliability is decided.

The failure we report is a property of that layer, and it has the shape that makes agent deployments
hard to trust. It is not an outage and not an empty field. The pipeline reports success, the text
looks complete, and the agent answers fluently and wrongly. Enterprise authors mark obligations
visually---a red requirement, a highlighted exception, a struck-through value that no longer
applies---and every mainstream converter we tested discards that layer while preserving the words
around it. An agent reading the result sees a list of requirements with no indication which ones the
author flagged, and, as we show, it guesses.

Three properties make this a safety-relevant ingestion problem rather than a formatting
inconvenience. First, it is \emph{silent}: nothing in the pipeline signals that information was
dropped, so no monitor fires. Second, it \emph{inverts} the usual failure mode: the agent does not
abstain or return less, it asserts more, and its false-positive rate on requirements the author did
\emph{not} flag rises fourfold. Third, it is \emph{common}, present in about a third of real
documents, so it is not an adversarial edge case but the ordinary state of enterprise corpora.

The setting is demanding in ways that compound this. Enterprise procedural documents span 50--300
pages and mix formal definitions, step-by-step procedures, compliance rules, exception clauses,
organisational charts, decision flowcharts, timelines and regulatory references---content whose
semantics a human reader partly reconstructs from presentation rather than from wording. Documents
are long enough that fixed-chunk ingestion fragments cross-references a reader resolves globally;
content is heterogeneous, and several kinds of it (headers, footnotes, tracked revisions) are
unreachable from a naive document walk; and authoring conventions are inconsistent, so the same
visual device means different things across departments and must be interpreted rather than mapped
one-to-one. Stated concretely, the problem is to produce a representation of such a document that a
model can read \emph{without losing what the author encoded visually}, and that is faithful, complete
and economical enough to fit a context window.

This paper aims to contribute on three fronts. We \textbf{quantify} what formatting-blind ingestion
costs an agent downstream, on real documents and against ground truth taken from the file's own
markup rather than from our judgement, and we show the effect holds across six answering models from two vendors. We \textbf{localise} it: a span-level leave-one-out over six cue classes
attributes essentially the whole effect to colour and highlighting, with bold, italic, monospace and
strikethrough contributing nothing measurable---which turns ``formatting matters'' into a specific,
actionable claim about the two cues no mainstream converter preserves and no parsing benchmark
scores. And we ask whether recovering a cue is enough, or whether an agent also needs to be told what
the cue \emph{means}; a pre-registered test says the latter, though the margin is modest and we
report it as such. Alongside these we release a validated evaluation instrument: labels derived from
OOXML run properties, scored deterministically with no LLM judge, checked against rendered page
images through an independent modality, and bounded above by a supervised text-only classifier so
that the task's cue-dependence is measured rather than asserted. We also report what did not work,
since both bear on how such systems should be built: progressive disclosure did not reduce token
cost, and the typed layer shows no benefit on the two models that attempt the task only sparsely.

\section{Related Work}
\label{sec:related}

\textbf{Document conversion and benchmarks.} A mature toolchain converts documents for agent
consumption: \texttt{python-docx} and \texttt{docx2python} read OOXML directly, \texttt{mammoth} maps
it to HTML, and Docling~\cite{docling2024}, Unstructured~\cite{unstructured} and \texttt{pymupdf4llm}
target LLM-ready Markdown. These are evaluated on text accuracy, reading order, tables and
formulas---the axes of OmniDocBench~\cite{omnidocbench2024} and ParseBench~\cite{parsebench2026}---so
character-level presentation sits largely outside that frame. Microsoft's Office Comprehension
Benchmark~\cite{ocb2026} is the first public suite evaluating LLM systems jointly on native Word,
Excel and PowerPoint files, and we evaluate on its Word File-Fidelity track directly
(\S\ref{sec:eval_ocb}); even there, formatting-dependent items are a minority and none turns on what a
surviving cue \emph{means}.

\textbf{Formatting recognition.} The closest refereed work is TexTAR~\cite{textar2025}, an ICDAR
2025 oral recognising textual attributes in document images. It covers precisely four---bold,
italic, underline, strikeout---and neither colour nor highlighting, the two our ablation finds
load-bearing. It also works from images and stops at attribute detection, where we recover cues
from OOXML and PDF geometry, assign semantic roles, and measure the effect on answer correctness.
Layout analysis~\cite{huang2022layoutlmv3,xu2020layoutlm} models page geometry, which is
complementary: our documents are text-heavy and their semantics sit in character properties.

\textbf{Rendered versus ingested document.} Liu et al.~\cite{liu2026canvas} mine the OOXML
specification for 21 ``evidence forks'' under which a file presents one view in Office and another
after extraction---\emph{plural ground truth}. Every fork runs in the direction \emph{extraction
reveals what the canvas hides}, and their six dimensions contain no run-level character formatting;
ours runs the other way. Yang et al.~\cite{eyes2026usenix} establish the premise adversarially at
USENIX Security: moderation systems ``operate primarily on tokenized text and often overlook visual
cues that humans naturally use'', a gap they exploit to reach over $86\%$ human recognition at under
$1\%$ detection across 13 deployed systems. We differ from both in studying ordinary documents, where
nobody crafted anything.

\textbf{Is emphasis already a machine signal?} Two findings bracket the question our ablation
answers. PASTA~\cite{pasta2024} steers attention post hoc, motivated by exactly our premise---that
writers ``leverage the subtleties of text style, such as bold and italics, to guide the attention of
readers''---reporting a $22\%$ average gain for LLAMA-7B. Against it, Dillitzer et
al.~\cite{attentioncase2026} report an \emph{attention--performance divergence}: emphasis redirects
attention, yet ``increased concentration does not inherently improve task accuracy''. Whether gains
come from cue \emph{salience} or cue \emph{typing} is exactly what \S\ref{sec:eval_typing}
pre-registers. Relatedly, Braun et al.~\cite{hiddenstructure2025} show prompt-side structure matters,
but manipulate formatting the \emph{prompt author} controls, not what the \emph{document author} chose.

\textbf{Typed extraction.} Xu et al.~\cite{xu2026zoes} discover entity attributes without a
predefined schema; we adopt their few-shot finding but replace mutual-dependency validation with
source-quote verification, since a guided schema makes hallucination the risk that matters. Generic
LLM extraction is competitive given suitable prompting~\cite{wei2023zeroshot};
REBEL~\cite{cabot2021rebel} and ERNIE~\cite{zhang2019ernie} extract triples for knowledge graphs,
and work on legal contracts~\cite{lippi2019claudette} shares the need for typed extraction but
treats semantics as linguistically encoded. To our knowledge no prior system treats character-level
formatting as a first-class extraction cue, and no benchmark scores whether a cue's meaning reached
the model.

\section{Formatting-Aware Parsing}
\label{sec:parsing}

The parser reads the \texttt{.docx} file using \texttt{python-docx} and produces a flat list of \texttt{RawBlock} objects, each carrying structured formatting annotations. This stage is entirely code-based, with zero LLM calls.

\subsection{Formatting as First-Class Signal}
\label{sec:formatting_signal}

Parsing is the subject of this paper; the typed extraction and package rendering that consume its
output are summarised in Appendix~\ref{app:arch}. The parser reads the \texttt{.docx} file with
\texttt{python-docx} and emits a flat sequence of blocks carrying structured formatting annotations,
entirely in code with zero LLM calls. Every contiguous run of text with identical formatting becomes
a \textsc{FormattingSpan}, an immutable record of 12 attributes; eleven high-confidence
formatting-to-semantics mappings then attach \textsc{SemanticRole} hints---red text or red highlight
to \textsc{critical\_warning}, any other visible highlight to \textsc{warning}, strikethrough to
\textsc{deprecated}, and so on (Table~\ref{tab:semantic_roles}, Appendix~\ref{app:stage1}). These are \emph{hints}, not ground
truth: we deliberately exclude medium-confidence mappings such as green text to status marker,
because a wrong hint biases an agent worse than no hint.

Blocks are serialized in a hybrid form. Markdown carries what a model has seen extensively in
pretraining (\texttt{**bold**}, \texttt{\~{}\~{}struck\~{}\~{}}); custom tags carry what Markdown
cannot express (\texttt{[COLOR:\#FF0000]}, \texttt{[HIGHLIGHT:YELLOW]}); and structural tags carry
document-level metadata a model cannot infer from text alone, such as which paragraphs belong to one
ordered list. Appendix~\ref{app:stage1} enumerates the tag set.

\subsection{Cue Disambiguation and Recovery}
\label{sec:disambiguation}

Cues are ambiguous and must be resolved before the text reaches an agent, and two of our rules were
corrected during this study after corpus measurement contradicted the initial design. Bold means
\textsc{key\_term} when only some spans of a block carry it and \textsc{implicit\_heading} when the
block is a standalone line under 200 characters and entirely bold. Colour is matched by channel
threshold rather than hex literal, which admits dark reds while excluding orange---though we note
that $99.7\%$ of red spans in the corpus are exactly \texttt{\#FF0000}, so the tolerance buys less
than expected. Highlight is typed by \emph{presence}, not hue: our initial rule typed only yellow,
and measurement showed cyan and light-grey highlighting of obligations is common, so any highlight
except \texttt{WHITE}/\texttt{NONE}---which render identically to unhighlighted text---now types as
\textsc{warning}. That single correction raised the share of author-marked paragraphs receiving a
salience role from $79.1\%$ to $97.6\%$. Roles attach at \emph{block} level while spans carry raw
inline markers, and that separation is what makes the salience-versus-typing ablation possible.

Several kinds of content are unreachable from a \texttt{w:body} walk and were silently absent from
our own representation until an external benchmark exposed it: headers and footers, footnotes,
document properties, OMML equations, and---most consequentially---tracked changes, where a deletion
is a \texttt{w:r} nested inside \texttt{w:del} carrying \texttt{w:delText}, so the \emph{accepted
revision}, the text actually in force, is discarded. Appendix~\ref{app:stage1} gives the full rule
set and the OOXML paths.

\section{Evaluation}
\label{sec:eval}

Our claim is that a formatting cue carries meaning that survives or is destroyed during
parsing, and that whether it survives changes whether a downstream model answers
correctly. Testing this requires a task whose ground truth is the cue itself rather
than our interpretation of it. We therefore evaluate on three levels: a real-document
task at scale built from OOXML run properties (\S\ref{sec:eval_task}--\ref{sec:eval_typing}),
an external benchmark whose questions we did not write (\S\ref{sec:eval_ocb}), and
and a controlled test that the advantage survives an identical downstream pipeline (\S\ref{sec:eval_system}).

All experiments use Claude~Opus~4.6 unless stated, temperature~0. Question
generation and second-opinion judging use different model families, so no model both
writes and answers a question. Scripts, item sets, per-document content hashes and
raw result JSON accompany the paper.

\subsection{Corpus, prevalence, and task}
\label{sec:eval_task}

We scanned \textbf{6{,}462} real \texttt{.docx} policy documents from a public corpus, named by content SHA-256, detecting cues at OOXML run level so nothing is inferred from wording. Walking the whole package, $32.8\%$ carry deliberate red text or a non-white highlight and $10.3\%$ carry one on a sentence expressing a modal obligation---the risk-bearing case; a further $2.6\%$ carry cues \emph{only} outside the body, where no baseline we test can see them (Appendix~\ref{app:prevalence_full}).

This supports a real-document task at scale. A policy document contains several
requirement-like sentences; the author marked some (red text or highlight) and left
others unmarked. We ask:

\begin{quote}
\itshape List every requirement in this document that the author has singled out for
special attention or elevated priority.
\end{quote}

Ground truth is the marked set, taken directly from OOXML run properties---never from
judgement. Scoring is set retrieval (precision, recall, F1) with a false-positive rate
measured against real \emph{unmarked} requirements from the same document, so a system
that lists everything is penalised. Matching is lexical and deterministic: \textbf{no
LLM judge is involved}. A gold requirement counts as retrieved when a run of four
consecutive opening content words appears in the answer. That rule is deliberately
conservative, and we verified it does not depress the numbers: only \emph{one} of the 332
gold requirements has fewer than four content words and is therefore unmatchable by
construction, giving every arm a recall ceiling of $0.997$, and no unmarked distractor is
unmatchable at all. Relaxing the rule to a three-word run or 60\% content-word overlap
would accept a further $4.8\%$ of unmatched pairs, but inspection shows most of those are
the looser rule wrongly matching a \emph{different} requirement that shares boilerplate
wording, so we keep the strict form. Construction requires at least one marked and at least three
unmarked requirements per document, and rejects any document whose marked sentences
are the only ones carrying urgency words. This yields \textbf{84 documents}
(332 marked, 4{,}192 unmarked distractors); 76--79 are scored in a given run, as a
document is used only when every arm produces a representation under the size cap, so
comparisons stay paired.

\textbf{Is the task solvable without formatting?} We measure this rather than assert
it. Fitting a supervised text-only classifier (TF-IDF word 1--2 grams plus character
3--5 grams, logistic regression, \textbf{grouped by document} so no fold shares a
document with its training split) to the marked/unmarked labels gives
ROC~AUC~$=0.664$. The two interpretable heuristics a zero-shot reader could actually
use sit at chance: an urgency-keyword rule scores $0.508$ and sentence length
$0.492$. So the wording carries a weak signal recoverable only with supervision on the
labels, and this $0.664$ is a \emph{ceiling} on any formatting-blind method.

\subsection{Formatting-blind conversion destroys answerability}
\label{sec:eval_main}

Every arm receives the same question over the same document and differs only in how
the document was parsed (Table~\ref{tab:flagship}).

\begin{table}[htbp]
  \centering
\caption{Marked-requirement set retrieval on real documents. Upper panel: all six arms, 76 documents, 3 repeats, 1{,}368 calls, paired per-document macro F1 against \textsc{ours}. Lower panel: the \textbf{pre-registered} confirmatory test of the typing claim, run once on 79 documents with 6 repeats (1{,}422 calls) and three arms only, so the correction family is two hypotheses. Both arms in H1 show the model identical inline cue markers over byte-identical content, differing only in whether a typed role token is attached. Tests are exact two-sided paired sign tests, Holm-corrected, with 10{,}000-resample bootstrap CIs over documents.}
\label{tab:flagship}
\footnotesize
\setlength{\tabcolsep}{3pt}
\begin{tabular}{lrrrrr}
\toprule
\textbf{Arm} & \textbf{micro} & \textbf{prec.} & \textbf{FP} & \textbf{$\Delta$ macro} & \textbf{$p_{\text{Holm}}$} \\
 & \textbf{F1} & & \textbf{rate} & \textbf{F1} & \\
\midrule
\textsc{ours} & \textbf{0.690} & 0.824 & 0.011 & --- & --- \\
\textsc{ours}, salience roles & 0.687 & 0.834 & 0.010 & $-0.005$ & 1.00 \\
\textsc{ours}, untyped markup & 0.605 & 0.811 & 0.010 & $+0.092$ & \textbf{0.025} \\
docx2python & 0.586 & 0.797 & 0.010 & $+0.110$ & \textbf{0.022} \\
docling & 0.150 & 0.196 & 0.043 & $+0.578$ & $<10^{-4}$ \\
plain text & 0.075 & 0.437 & 0.005 & $+0.689$ & $<10^{-4}$ \\
\midrule
\multicolumn{6}{l}{\textit{Pre-registered confirmatory test} (79 docs, 6 repeats)} \\
\textsc{ours}, salience roles & \textbf{0.667} & 0.836 & 0.010 & --- & --- \\
\textsc{ours}, untyped markup & 0.597 & 0.818 & 0.009 & $+0.085$ & \textbf{0.049} \\
docx2python & 0.570 & 0.799 & 0.010 & $+0.107$ & \textbf{0.014} \\
\multicolumn{6}{l}{\footnotesize H1 typed $>$ untyped: 13/4/62, CI $[+.035,+.145]$} \\
\multicolumn{6}{l}{\footnotesize H2 typed $>$ docx2python: 17/4/58, CI $[+.047,+.174]$} \\
\bottomrule
\end{tabular}
\end{table}

Plain text scores \textbf{0.075} and docling \textbf{0.150}, against \textbf{0.690}
for our representation---a gap of $+0.689$ and $+0.578$ in per-document macro F1
($p<10^{-4}$, 62 of 76 and 57 of 76 documents won). These are not marginal effects.
A formatting-blind converter sees the same sentences with no way to tell which the
author marked, and the resulting answers are not merely incomplete: docling's
false-positive rate on genuinely unmarked requirements is $0.043$, roughly four times
ours, because a system that cannot see the cue guesses.

The comparison that isolates \emph{preservation} from \emph{pipeline depth} is
docx2python, which preserves character formatting as HTML and is the strongest
baseline throughout. It reaches $0.586$---far above docling, far below ours, and the
$+0.110$ gap to ours is significant even within this six-arm family
($p_{\text{Holm}}=0.022$).

\subsection{The loss is concentrated in colour and highlight}
\label{sec:eval_percue}

\begin{figure}[htbp]
    \centering
    \includegraphics[width=0.62\textwidth]{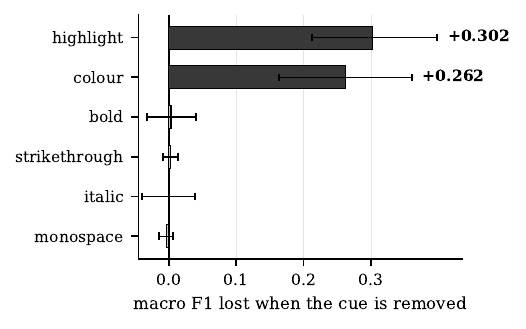}
    \caption{Loss in per-document macro F1 when exactly one cue class is removed at the span level, with 10{,}000-resample bootstrap CIs. Dark bars are significant after Holm correction. Colour and highlight account for essentially the entire effect; the other four cues sit within $\pm0.003$ of zero with intervals straddling it.}
    \label{fig:percue}
\end{figure}

``Formatting matters'' is too coarse a claim to be useful. We localise it with a
per-cue leave-one-out ablation: each arm removes exactly one cue class at the
\emph{span} level and re-derives the semantic roles, so document content is
byte-identical across arms and only that cue's markers and role token are absent
(Figure~\ref{fig:percue}).

The entire effect is carried by \textbf{highlight} ($+0.302$,
$p_{\text{Holm}}<10^{-5}$) and \textbf{colour} ($+0.262$, $p_{\text{Holm}}=2\times10^{-5}$).
Bold, italic, monospace and strikethrough contribute nothing measurable---each within
$\pm0.003$ of zero, with confidence intervals straddling it. This is not because those cues
are absent: a single document carries over a thousand bold markers. It is because they are
too \emph{common} to be informative. An author who bolds every heading and defined term has
not thereby singled anything out, whereas colouring or highlighting a requirement is a
deliberate, sparing act. Cue typing pays only for cues rare enough to signal intent.

The ablation is also \emph{cue-specific}, which validates that the manipulation does what it
claims. Removing colour collapses colour-marked documents from $0.699$ to $0.214$ micro F1
while leaving highlight-marked documents at $0.694$; removing highlight does the mirror image,
$0.659 \to 0.128$, with colour-marked documents unchanged at $0.670$. Each removal therefore
damages exactly the documents that use that cue and moves the others by at most $0.011$.

This localisation matters because colour is precisely the cue that mainstream
conversion tools drop and that parsing benchmarks do not score. ParseBench's
formatting rules cover bold, italic, superscript and strikeout presence, with no rule
type for font colour at all.

\subsection{Does typing add anything beyond preserving?}
\label{sec:eval_typing}

The novel component of this system is not preserving cues---docx2python does that---but
attaching typed roles to them. Isolating that contribution requires comparing our
representation against \emph{the same representation with the role tokens removed and
every inline cue marker retained}.

This effect proved unstable across exploratory runs, for a reason we can name.
Recovering headers, footers, footnotes and document properties (\S\ref{sec:eval_ocb})
raised the untyped arm from $0.575$ to $0.610$, shrinking the gap, and enlarging the
arm family made the Holm correction stricter. Only 16 of 76 document pairs were
discordant, so the test was underpowered either way. Rather than re-analyse the same
data with different arm families, we \textbf{pre-registered} a confirmatory design---
hypotheses, three arms, six repeats, test and correction family fixed in advance and
committed to version control before execution.

We report that protocol's execution history in full, because the borderline $p$ below
depends on it. The design was executed once, then executed a second time, unchanged,
after a defect in our parser was fixed: a table-cell renderer was emitting neutral colour
markers (\texttt{[COLOR:\#000000]} on ordinary black body text) which are spurious cues,
so every \textsc{ours} representation changed and the earlier numbers could not stand.
Nothing about the hypotheses, arms, repeats, test or correction family was altered. The
pre-fix run gave $+0.084$ with W/L/T $15/2/62$ and $p_{\text{Holm}}=0.005$; the post-fix
run, which is what we report, gives $+0.085$ with $13/4/62$ and $p_{\text{Holm}}=0.049$.
The correction therefore made our headline $p$ \emph{worse} while leaving the effect size
unchanged, and the conclusion survives a change to the very parser it measures.

Both hypotheses are supported (Table~\ref{tab:flagship}, lower panel). Typing a preserved
cue adds $+0.085$ macro F1 over showing the identical cue untyped (CI $[+0.035,+0.145]$,
$p_{\text{Holm}}=0.049$), and our representation beats the strongest preserving baseline by
$+0.107$ (CI $[+0.047,+0.174]$, $p_{\text{Holm}}=0.014$). Six repeats rather than three was
decisive: averaging over more repeats removes per-call noise from each document's score
instead of parking it in the tie bucket, turning a 3/13 discordance against us into 13/4 in
favour. H1 sits close to the threshold and we report it as such---the effect is consistent
in sign and magnitude across every configuration we ran, but $+0.085$ should be read as an
estimate with a wide interval rather than a precise quantity.

The typed arm here is the \emph{salience-filtered} vocabulary---retaining only
\textsc{critical\_warning}, \textsc{warning} and \textsc{deprecated}---rather than the
full role set, for a reason established in \S\ref{sec:eval_general}.

\subsection{External validation: Microsoft OCB}
\label{sec:eval_ocb}

The preceding task is ours, so we also evaluate on the File-Fidelity Word track of Microsoft's
Office Comprehension Benchmark~\cite{ocb2026}: \textbf{326 items over 59 documents across 14
features}, with questions written by Microsoft and scored by OCB's own weighted assertion
satisfaction. Results are in Table~\ref{tab:ocb}.

\begin{table}[htbp]
  \centering
\caption{Microsoft OCB, File-Fidelity Word track: mean weighted assertion score over 326 items in 59 documents (1{,}630 scored rows, none dropped), with questions written by Microsoft. The 5 features with $n\!\leq\!7$ are aggregated; the full 14-feature breakdown is in Appendix~\ref{app:ocb}. \textbf{Bold} marks the best system per row. Lower block: paired over items against ours, exact sign test, Holm-corrected.}
\label{tab:ocb}
\small
\begin{tabular}{lrrrrrr}
\toprule
\textbf{Feature} & $n$ & \textbf{plain} & \textbf{docling} & \textbf{unstr.} & \textbf{dx2py} & \textbf{ours} \\
\midrule
Table & 143 & 0.916 & 0.939 & 0.932 & \textbf{0.986} & \textbf{0.986} \\
Text & 59 & 0.977 & 0.977 & 0.977 & 0.977 & \textbf{1.000} \\
Font & 24 & 0.083 & 0.312 & 0.292 & 0.604 & \textbf{0.750} \\
Hyperlink & 16 & 0.062 & \textbf{1.000} & 0.062 & \textbf{1.000} & \textbf{1.000} \\
Layout & 14 & 0.357 & 0.196 & 0.071 & 0.714 & \textbf{0.821} \\
Metadata & 14 & 0.000 & 0.000 & 0.000 & 0.000 & \textbf{1.000} \\
Comment & 11 & 0.091 & 0.091 & 0.091 & 0.091 & \textbf{1.000} \\
Highlight & 9 & 0.333 & 0.333 & 0.333 & \textbf{1.000} & 0.889 \\
Footnote & 8 & 0.000 & 0.000 & 0.000 & 0.000 & \textbf{1.000} \\
other (5 features) & 28 & 0.214 & 0.446 & 0.500 & 0.536 & \textbf{0.929} \\
\midrule
\textit{formatting} & 38 & 0.184 & 0.408 & 0.263 & 0.750 & \textbf{0.816} \\
\textit{non-formatting} & 288 & 0.693 & 0.761 & 0.722 & 0.818 & \textbf{0.977} \\
\textbf{all items} & 326 & 0.634 & 0.720 & 0.668 & 0.810 & \textbf{0.959} \\
\midrule
$\Delta$ vs.\ ours & & $+.325$ & $+.239$ & $+.290$ & $+.148$ & --- \\
$p_{\text{Holm}}$ & & \multicolumn{4}{c}{all $<10^{-6}$} & --- \\
\bottomrule
\end{tabular}
\end{table}

We lead on the 38 formatting-dependent items ($0.816$ against $0.750$) \emph{and} on the 288 that are
not about formatting ($0.977$ against $0.818$), so the gain is not confined to the dimension this paper
argues about. We are also the only system evaluated that
answers document-property, comment and footnote questions at all.

Running the full track against our own system exposed two defects, both since fixed. We scored \textbf{0.000} on Header and Footer items where \texttt{unstructured} scored $1.000$, and every system including ours scored $0.000$ on Metadata and Footnote: none of headers, footers, footnotes, document properties or OMML equations is reachable from a \texttt{w:body} walk, so all five were silently absent. Adding them raised our overall score from $0.853$ to $0.959$. We also report where we lose: one Highlight item to docx2python, and a tie with docling on Equation.

\subsection{Generalisation, and what does not replicate}
\label{sec:eval_general}

\begin{figure}[htbp]
    \centering
    \includegraphics[width=\textwidth]{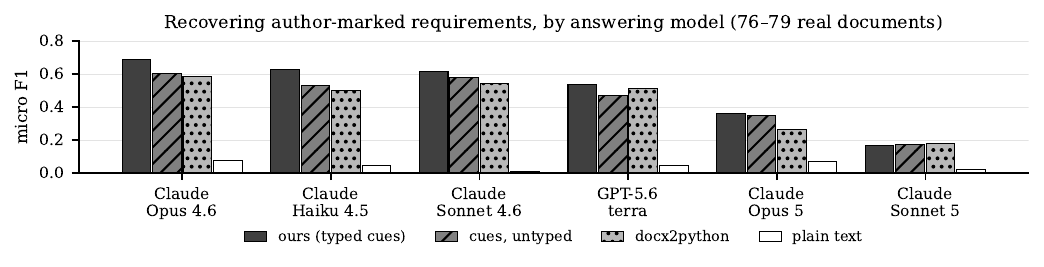}
    \caption{Recovery of author-marked requirements under six answering models spanning two vendors and three capability tiers. Plain text collapses under every model ($0.007$--$0.075$), so the failure is a property of the representation and not of a particular model. docling is omitted because it was not run under every model; it appears for all arms in Table~\ref{tab:flagship}.}
    \label{fig:cross_model}
\end{figure}

\textbf{A second format.} The same asymmetry appears in PDF, which has no run properties, so cues
must be recovered geometrically from character bounding boxes. Over all 126 born-digital formatting
rules of ParseBench we match the strongest PDF baseline exactly on cue recovery ($65/126$ each) while
plain text extraction recovers \emph{none} ($0/126$) despite locating $107$ of $126$ targets. The only
per-cue difference distinguishable in either direction is highlighting, where we lead $11/18$ against
$4/18$ (exact McNemar, $p=0.039$). Table~\ref{tab:pdf} in Appendix~\ref{app:pdf} gives the breakdown
and the OCR caveat.

\textbf{Six answering models.} Repeating the task across two vendors and three capability tiers
separates a robust finding from a fragile one (Figure~\ref{fig:cross_model}; per-model statistics in
Table~\ref{app:tab:crossmodel}). Cue \emph{preservation} replicates overwhelmingly: plain text
collapses under every model, from $0.007$ to $0.075$ micro F1. Beating docx2python replicates on four
of six. Cue \emph{typing} replicates on three of six, is positive on five, and is never significantly
negative---but it is not universal, and we say so.

\textbf{Where it fails.} The two models showing no typing benefit are the two that barely attempt the task: Claude Sonnet~5 answers at recall $0.09$ with precision $0.96$. The same missing cue therefore produces \emph{opposite} pathologies depending on the model---over-flagging in some, extreme abstention in others---and neither is detectable without ground truth.

The typed arm we report is the \emph{salience-filtered} vocabulary, retaining only
\textsc{critical\_warning}, \textsc{warning} and \textsc{deprecated}. The full role set is dominated by
structural tokens: across 40 documents \textsc{implicit\_heading} occurs 2{,}136 times and
\textsc{emphasis} 1{,}338, against 1{,}004 for \textsc{critical\_warning} and 452 for
\textsc{warning}. Roughly $62\%$ of role tokens therefore say nothing about whether the author singled
a passage out, and a model that treats any role token as a flag over-triggers---which is what we
observed on a weaker model excluded by the competence screen below. Restricting the vocabulary fixes
it, and under Claude the filtered and full sets are statistically indistinguishable ($p=0.45$).

\textbf{Model eligibility was decided by a criterion, not by inspection.} Typed hints require a model
to use a hint \emph{discriminatively}, which is an instruction-following capability, so a model that
cannot do the task at all cannot be interpreted on formatting cues. Each candidate therefore answered
the same question over a \emph{formatting-free} control document in which the flagged requirements are
named explicitly in prose, with thresholds fixed in advance (macro F1 $\geq 0.80$, false-positive rate
$\leq 0.05$). All six reported models pass (macro F1 $0.874$--$0.976$, FP rate $0.003$--$0.037$). One
further model, Mistral~Large~3, fails both at $0.558$ and $0.133$: its control \emph{precision} is
$0.480$, meaning it flags about as many unmarked requirements as marked ones \emph{when the answer is
written out in words}. Over-flagging is a baseline property of that model rather than a response to
our hints, so it is out of population; we report its numbers in Appendix~\ref{app:crossmodel} rather
than discarding them.

\subsection{Does the advantage survive an identical downstream pipeline?}
\label{sec:eval_system}

Handing a parsed representation straight to the answering model invites the objection that we
compared a full pipeline against bare converters. Running the \emph{complete} pipeline per arm
over 68 documents---an identical extraction prompt, model and tier logic, identical image bytes,
identical renderer, varying only the parsed text---the advantage survives and grows (Table~\ref{tab:equalprep}). \begin{table}[t]
  \centering
\caption{Full pipeline run per arm on 68 documents, 2 repeats, 174 packages built and 408 scored calls. Every arm receives the same extraction and rendering treatment; only the parsed text differs. ``retention'' is the share of gold requirements whose text is still present anywhere in the rendered package, and is near-identical across arms.}
\label{tab:equalprep}
\footnotesize
\setlength{\tabcolsep}{4pt}
\begin{tabular}{lrrrrr}
\toprule
\textbf{Parser} & \textbf{micro F1} & \textbf{macro F1} & \textbf{retention} & \textbf{$\Delta$} & \textbf{$p_{\text{Holm}}$} \\
\midrule
\textsc{ours} & \textbf{0.503} & \textbf{0.526} & 0.665 & --- & --- \\
docx2python & 0.258 & 0.347 & 0.634 & $+0.178$ & \textbf{0.0001} \\
docling & 0.084 & 0.104 & 0.607 & $+0.422$ & $<10^{-4}$ \\
\bottomrule
\end{tabular}
\end{table}

The interpretive key is that all three arms retain a \emph{similar} share
of the gold requirement text in the final package ($0.665$, $0.634$, $0.607$), so the gap is not
content loss: only our package still records \emph{which} requirements the author singled out.
Table~\ref{tab:equalprep} gives the numbers.

\subsection{Implications for agents deployed in the wild}
\label{sec:eval_agents}

Three consequences follow for anyone running document-consuming agents in production.

\textbf{The failure is invisible to standard monitoring.} Every arm we tested returns
well-formed output on every document; none raises, none reports missing content. A pipeline
health check, a schema validator, and a token-count sanity check all pass. The only signal is
answer quality on questions whose ground truth lives in the source markup---which is precisely
what an operator does not have. This argues for ingestion-level assertions rather than
output-level monitoring: our own parser now asserts that a recovered cue implies a located span,
and that assertion caught a defect we had already published in a table.

\textbf{It degrades precision, not just recall, which is the wrong direction for autonomy.} An
agent that returns less is inconvenient; an agent that confidently returns \emph{more} is
dangerous, because a downstream action may be taken on a requirement the author never flagged.
docling's false-positive rate on unmarked requirements is $0.043$ against our $0.011$. Under two
of the six answering models the effect appears as extreme conservatism instead---recall $0.09$ at
precision $0.96$---so the same missing signal produces opposite pathologies depending on the
model, and neither is detectable without ground truth.

\textbf{The fix is cheap, and it belongs before the agent.} Recovering these cues is
deterministic, code-only, and costs no LLM calls; the entire effect reported here comes from
parsing alone. It does cost tokens---roughly twice plain text---so it is a deliberate trade of context
budget for factuality rather than a free win. Notably, the cue must be \emph{typed} and not merely
passed through: showing an agent \texttt{[COLOR:\#FF0000]} is worth less than telling it the span
is a critical warning ($+0.085$, $p_{\text{Holm}}=0.049$), which is consistent with evidence that
emphasis redirects model attention without accuracy reliably following~\cite{attentioncase2026}.

\section{Conclusion}
\label{sec:conclusion}

An enterprise agent reads whatever its ingestion layer produces, and that layer silently discards
the meaning authors encode visually. On 76 real policy documents, recovering the requirements an
author singled out falls from micro F1 $0.690$ to $0.150$ (docling) and $0.075$ (plain text), a gap that replicates across six answering models from two vendors. The failure is the
difficult kind for
autonomy: not an error the pipeline reports, but an agent that answers fluently and wrongly, with
four times the false-positive rate on requirements the author never flagged.

The loss is localised. A per-cue leave-one-out attributes essentially all of it to
\textbf{highlight} ($+0.302$) and \textbf{colour} ($+0.262$), both $p_{\text{Holm}}<10^{-4}$, with
bold, italic, monospace and strikethrough each within $\pm0.003$ of zero---and the same asymmetry
reappears in PDF. Colour is simultaneously the cue that matters most and the one no mainstream
converter preserves and no parsing benchmark scores. A pre-registered test further shows that
\emph{typing} a cue for the agent adds $+0.085$ beyond preserving it
($p_{\text{Holm}}=0.049$), consistent with evidence that emphasis redirects attention without
accuracy reliably following~\cite{attentioncase2026}.

\textbf{Limitations.} Exposure is common but not universal: a third of documents carry a deliberate
colour or highlight and $10.3\%$ carry one on an obligation. PDF support requires a usable text layer,
since we do not OCR. The typed layer is robust on three of the six models but shows no benefit on two that attempt the task only sparsely, so it is conditional on the model engaging with the hint. Ground
truth is derived automatically from run properties and validated against rendered pages by an
independent modality ($\kappa=0.72$) but not yet by human annotators, which is the most valuable
addition remaining. The task is unsaturated at recall $0.60$, and this is genuine difficulty rather
than measurement: the scorer's structural ceiling is $0.997$. Our representation costs roughly twice
plain text in tokens, a deliberate trade of context budget for factuality. Absolute micro F1 is stable
only to about $\pm0.08$ across configurations, so every claim rests on paired per-document tests.

\textbf{Future work.} Human validation of the labels; OCR-backed cue recovery for scanned pages;
non-English documents, where these cues carry culturally inflected semantics; and diff-based
re-ingestion across revisions, for which tracked changes already supply reviewer-signed ground truth.

\appendix
\setcounter{section}{0}
\renewcommand{\thesection}{S\arabic{section}}

\section{Corpus and Prevalence}
\label{app:prevalence_full}

\label{sec:eval_corpus}

We scanned \textbf{6{,}462} real \texttt{.docx} files from a public corpus of
policy and procedural documents (files are named by content SHA-256, so provenance
is verifiable and de-duplication is exact). Scanning is done at OOXML run level, so
nothing is inferred from wording. Appendix~\ref{app:prevalence} gives the full
breakdown by pattern: a colour or highlight marking on a constraint occurs in 359
documents, manual strikethrough in 35, and Word tracked changes in 40. Tracked
changes were invisible to our earlier scans, because a tracked deletion is a
\texttt{w:r} nested inside \texttt{w:del} carrying \texttt{w:delText}, which
neither \texttt{paragraph.runs} nor \texttt{iter\_inner\_content} reaches.

Two figures matter here and they differ by a factor of three, so we give both.
Walking the entire OOXML package, \textbf{32.8\%} of documents carry deliberate red
text or a non-white highlight somewhere, and \textbf{10.3\%} carry one on a sentence
expressing a modal obligation---the risk-bearing case, since that is where a dropped
cue changes what a reader is told to do. The stricter pattern~C above (5.6\%) requires
the marked \emph{run} rather than the marked \emph{sentence} to carry the obligation,
and therefore undercounts by roughly $1.8\times$; we report it because it is the
population the task in \S\ref{sec:eval_task} is built from, not because it is the
prevalence of the phenomenon. A further \textbf{2.6\%} of documents carry cues
\emph{only} outside the body---in headers, footers, footnotes or comments---which no
baseline we test can see at all.

\section{OCB Per-Feature Results}
\label{app:ocb}

\section{Tracked Changes}
\label{app:revision}

\textbf{Tracked changes.} Word revisions give the strongest ground truth available---the reviewer's own
recorded edit. Across 28 recorded revisions ours carries both the superseded and in-force value with
the superseded one \emph{marked} in 27 cases, while no other arm marks it even once; plain text loses
the value entirely in 16 and surfaces \emph{only the superseded} value in 3, an outright factual error.
On forced-choice accuracy we and docx2python are indistinguishable ($0.818$ versus $0.833$ at
$n\!\approx\!35$), because a converter that silently discards \texttt{w:del} is \emph{accidentally
right}. At $n=12$ items this cannot carry a claim and we do not rest one on it.

\begin{table}[t]
  \centering
\caption{Semantic role assignments. Only HIGH-confidence formatting-to-semantics mappings are included. Medium-confidence rules (e.g., italic $\rightarrow$ document reference, green text $\rightarrow$ status marker) are excluded to avoid misleading the extraction model with ambiguous signals. ``Visible highlight'' means any hue other than \texttt{WHITE}/\texttt{NONE}, which render identically to unhighlighted text (\S\ref{sec:disambiguation}); ``standalone'' means a single line under 200 characters. Appendix~\ref{app:formatting_pipeline} gives the full parse-time disambiguation rule, serialization, and extraction signal for each role.}
\label{tab:semantic_roles}
\small
\begin{tabular}{lll}
\toprule
\textbf{Formatting} & \textbf{Semantic Role} & \textbf{Signal} \\
\midrule
Bold, some spans only & \textsc{Key\_Term} & Key terms \\
Bold, standalone block & \textsc{Implicit\_Heading} & Section heads \\
Italic/underline, not all-bold & \textsc{Emphasis} & Emphasis \\
Red text or red highlight & \textsc{Critical\_Warning} & Critical alerts \\
Any other visible highlight & \textsc{Warning} & Cautions \\
Strikethrough & \textsc{Deprecated} & Obsolete content \\
Monospace font & \textsc{Code\_or\_Literal} & Code/literals \\
Boxed/bordered & \textsc{Callout} & Callout boxes \\
Superscript & \textsc{Reference\_Marker} & Footnotes \\
ALL CAPS line & \textsc{Implicit\_Heading} & Section heads \\
Header + DRAFT & \textsc{Metadata} & Doc metadata \\
\bottomrule
\end{tabular}
\end{table}

\section{Stage-1 Rules in Full}
\label{app:stage1}

\subsection{Formatting Disambiguation}
\label{sec:disambiguation}

The raw mapping in Table~\ref{tab:semantic_roles} conceals ambiguity that must be resolved
before the text reaches a model. The rules below are the ones the released parser applies;
two of them were corrected during this study after corpus measurement contradicted our
initial design, and we give both the rule and the correction because the corrections are
themselves findings.

\textbf{Inline versus standalone bold.} Bold triggers \textsc{key\_term} or
\textsc{implicit\_heading} depending on context, and the discriminator is \emph{not} a
coverage ratio. A block is \emph{standalone} when its plain text is a single line shorter
than 200 characters. A standalone block whose every span is bold---or whose text is
entirely upper case---becomes \textsc{implicit\_heading}; otherwise a block with some but
not all spans bold becomes \textsc{key\_term}. One consequence is worth stating because it
is a genuine gap rather than a design choice: an all-bold paragraph of 200 characters or
more matches neither branch and receives no bold-derived role at all (29 such blocks in a
29,170-block sweep). We considered a character-coverage ratio instead and found it no
better on the threshold sweep, so the simpler standalone test ships.

\textbf{Colour as fuzzy matching.} Authors do not use canonical RGB values: a red warning
may be \texttt{\#FF0000}, \texttt{\#C00000}, \texttt{\#CC0000} or a custom hue chosen years
ago. Rather than matching hex literals the parser thresholds the channels,
\textit{is\_red}$(r,g,b) = (r>180) \wedge (g<80) \wedge (b<80)$, which admits dark reds
while excluding orange and maroon. We report one honest deflation of this design: on the
real corpus $99.7\%$ of red spans are exactly \texttt{\#FF0000}, so the tolerance costs
nothing but buys much less than we expected.

\textbf{Highlight is typed by presence, not by hue.} Our initial rule typed only yellow
and dark-yellow highlights as \textsc{warning}. Corpus measurement showed this was wrong:
cyan and light-grey highlighting of obligations is common, and those spans reached the
model carrying their raw value with no role attached. The shipped rule therefore types
\emph{any} highlight as \textsc{warning} except \texttt{WHITE} and \texttt{NONE}, which
render identically to unhighlighted text and are not marks at all; a \texttt{RED} highlight
escalates to \textsc{critical\_warning} alongside red text. Hue is still preserved in the
serialized output, so a reader that wants to distinguish yellow from cyan can. This change
raised the share of author-marked paragraphs receiving a salience role from $79.1\%$ to
$97.6\%$, and it is the single correction with the largest effect on our results.

\textbf{Roles are attached at block level.} A role is computed from the properties of a
block's spans but recorded on the block, serialized as a tag attribute
(\texttt{roles="CRITICAL\_WARNING"}). Individual spans carry their raw cue markers inline
(\texttt{[COLOR:\#FF0000]}, \texttt{[HIGHLIGHT:YELLOW]}, \texttt{\textasciitilde\textasciitilde
struck\textasciitilde\textasciitilde}) but not role names. This distinction is what makes
the ablation in \S\ref{sec:eval_typing} possible: removing the \texttt{roles} attribute
leaves every inline marker intact, so cue \emph{salience} and cue \emph{typing} can be
varied independently over byte-identical content.

\subsection{Hybrid Serialization Protocol}
\label{sec:serialization}

After parsing, blocks are serialized to an annotated text string for LLM consumption. The serialization design involves three interlocking choices:

\textbf{Hybrid Markdown + custom tags.} Standard Markdown handles formatting the LLM has seen extensively in pretraining: bold (\texttt{**text**}), italic (\texttt{*text*}), and strikethrough (\texttt{\~{}\~{}text\~{}\~{}}). For formatting with no Markdown equivalent, we use custom tags: \texttt{[COLOR:red]...[/COLOR]}, \texttt{[HIGHLIGHT:yellow]...[/HIGHLIGHT]}. This hybrid is 15--20\% more token-efficient than pure custom-tag serialization while exploiting the model's Markdown familiarity.

\textbf{Structural metadata tags.} Every block is wrapped in a structural tag (e.g.\ \texttt{[HEADING level=2 style="Heading 2"]}, \texttt{[TABLE has\_header=True cols=4]}) carrying document-level metadata that the LLM cannot infer from text alone: which numbered paragraphs are part of an ordered list, which table has a header row, which paragraph sits under a given heading level. Without these, the extraction model would need to re-derive heading hierarchy from font size heuristics, a fragile and error-prone operation. Tag attributes additionally carry fine-grained context---indentation depth, font size, and Word style name. Appendix~\ref{app:serialization_details} enumerates the full tag set and attribute encoding.

Appendix~\ref{app:formatting_pipeline} details the complete formatting-to-LLM pipeline for all 11 active SemanticRoles: the raw Word attribute, the disambiguation rule applied at parse time, the resulting role, the exact serialization the LLM receives, and the extraction signal that role encodes. Appendix~\ref{app:parsing_example} walks a worked parsing example end to end, and Appendix~\ref{app:docx_limits} documents the XML-level workarounds required for content \texttt{python-docx} does not natively expose (text boxes, tracked changes, comments, merged cells, OLE objects).

\subsection{Content Unreachable from a Body Walk}
\label{sec:outofbody}

A \texttt{.docx} is an OPC package, not a single stream, and several kinds of content sit
outside the part that a naive walk traverses. Each of the following was silently absent
from our own representation until an external benchmark exposed it
(\S\ref{sec:eval_ocb}), and each remains absent from most converters we test:

\begin{itemize}[leftmargin=*,nosep]
  \item \textbf{Headers and footers} live in separate parts, up to three per section
  (default, first page, even page). They carry document-control facts---version, effective
  date, classification marking---and a red \textsc{draft} or highlighted classification in a
  header is precisely the signal this work concerns, so we parse them with character
  formatting intact and label which is primary.
  \item \textbf{Footnotes, endnotes and document properties} occupy their own parts;
  Word's separator notes must be skipped.
  \item \textbf{Tracked changes.} A tracked deletion is a \texttt{w:r} nested inside
  \texttt{w:del} carrying \texttt{w:delText} rather than \texttt{w:t}, and an insertion nests
  inside \texttt{w:ins}. Neither is a direct child of the paragraph, so both are invisible to
  a run-level walk---which means the \emph{accepted revision}, the text actually in force, is
  discarded. We walk the paragraph's children in document order and map the semantics onto
  the existing vocabulary: an insertion is current text, a deletion is content the author
  retired, which is what strikethrough already means here.
  \item \textbf{Equations, hyperlinks and text boxes}, whose characters sit in
  \texttt{m:t} inside \texttt{m:oMath} or in runs nested rather than direct.
\end{itemize}

Two robustness points follow from running at corpus scale. Real documents violate the
specification---one declares a fractional half-point font size, which makes a strict
reader raise and abandon the whole file---so a malformed cue must degrade to ``cue
absent'', never to ``document unreadable''. And python-docx reconstructs each table row
from the table grid, raising when a vertical merge references a row that supplies no cell
at that offset; we fall back to reading the row's cells directly from the XML, which loses
merge continuation but preserves the text. Both defects aborted a 6{,}462-document scan on
first contact.

For born-digital PDF the same cues must be recovered geometrically, since PDF has no run
properties: a highlight is a filled rectangle drawn behind glyphs, and a strikethrough is
a hairline crossing the glyph midline where an underline sits at the baseline. We
therefore classify from character-level bounding boxes rather than span-level ones---a
span-level comparison fails because the renderer merges struck and live text into a single
span (\S\ref{sec:eval_pdf}).

\section{System Architecture and Tiering}
\label{app:arch}
\label{sec:architecture}

\begin{figure}[h]
    \centering
    \includegraphics[width=\textwidth, keepaspectratio]{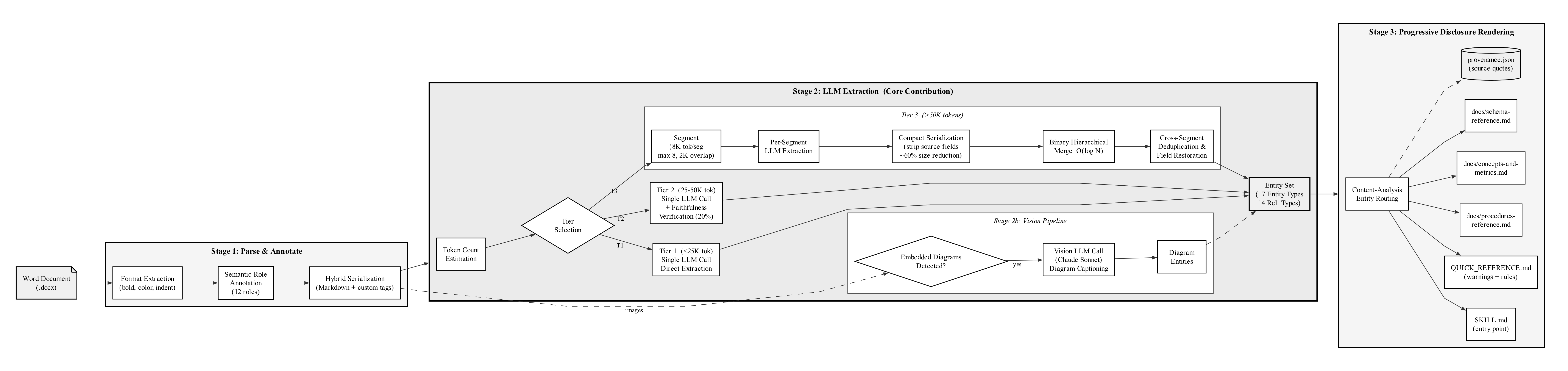}
    \caption{Three-stage pipeline. \textbf{Stage~1} (Parse \& Annotate, \emph{the focus of this paper}) recovers character-level formatting---including content unreachable from a body walk, such as headers, footers, footnotes, document properties, tracked changes and equations---and serializes it into a hybrid Markdown+tag format. \textbf{Stage~2} auto-selects a tier: a single call under 25K tokens, one call plus faithfulness verification to 50K, and above that segmentation with binary hierarchical merge at $O(\log N)$ depth, costing $2N{-}1$ calls for $N\leq16$ segments; a parallel vision pass (dashed) handles embedded diagrams. \textbf{Stage~3} routes the resulting typed entities into a five-file progressive-disclosure package. Stages~2--3 are reported as downstream utility (\S\ref{sec:downstream}), not as the object of measurement.}
    \label{fig:pipeline}
\end{figure}

The system processes documents in three stages (diagrammed in Figure~\ref{fig:pipeline}, Appendix~\ref{app:principles}). \textbf{Stage~1
(Parse \& Serialize)} is entirely code-based with zero LLM calls, and is the subject of
this paper: it recovers character-level presentation and emits annotated text. Stage~2
(Extract) and Stage~3 (Render) consume that text and are described in
\S\ref{sec:downstream}. The separation concentrates LLM reasoning where it adds value---
entity classification, relationship identification, editorial summarization---and keeps
deterministic work in code. Five design principles guide the architecture
(Appendix~\ref{app:principles}), the most consequential being \emph{formatting is signal,
not decoration} and \emph{source-quote provenance}.

\section{Adaptive Tiering by Document Size}
\label{sec:tiers}

Because Stage~2's cost is what makes a formatting-faithful representation practical at
enterprise document sizes, we note the tiering here and specify it in
Appendix~\ref{app:merge}. A tier is selected from the annotated document's estimated token
count: a single extraction call below 25K tokens (most enterprise SOPs), the same plus
faithfulness validation on a 20\% sample to 50K, and above that segmentation into up to 16
overlapping $\sim$8K-token chunks merged pairwise, costing $2N{-}1$ calls for $N$ segments
--- 3 at $N{=}2$, rising to 31 at the maximum. Segments are kept small deliberately so each
call completes inside the service's $\sim$330\,s streaming limit; wall-clock time is 30 seconds
to 8 minutes depending on tier.

A uniform 4K-token chunking alternative would need an estimated 40--80 calls for a
100-page document and would fragment extractions across chunk boundaries; that figure is
analytical, from the chunk count, and not a measured baseline. The tiered design instead
exploits the fact that one long-context call yields globally coherent output, able to
deduplicate and resolve cross-references in a single pass.

\section{From Representation to Agent-Loadable Knowledge}
\label{app:downstream}
\label{sec:downstream}

A faithful representation is a means, not an end. Our reference implementation carries it through
two further stages so an agent can load the result directly: one LLM call per document (or per
segment) converts the annotated text into 17 typed entity types joined by 14 relationships under a
mandate that every entity carry a verbatim \texttt{source\_quote}, and a content-analysis router
then writes those entities into a five-file progressive-disclosure package in which the entry
point holds operative guardrails and reference material moves to sub-files
(Appendices~\ref{app:pipeline}--\ref{app:disclosure}). Document size is handled by tiering
(\S\ref{sec:tiers}); embedded diagrams add one vision call per non-decorative image.

What matters for an agent is that the cue survives the whole pipeline. Measured over 34 generated
packages, a colour- or highlight-marked requirement reaches the quick-reference file at rate
$0.569$ against $0.245$ for an unmarked one---a $2.3\times$ enrichment---so an agent that loads
only the entry point and the cheat sheet still encounters disproportionately what the author
flagged. \S\ref{sec:eval_system} tests whether the Stage-1 advantage survives this pipeline;
generalising it to a measured claim about typed-extraction accuracy would need a benchmark that
does not yet exist, and is future work.

\begin{table}[htbp]
  \centering
\caption{All 126 born-digital formatting rules of ParseBench. Cells are \emph{cue recovered\,/\,text located}; cue $\leq$ text necessarily. Plain text locates almost every target and recovers no cue at all. $p$ is an exact McNemar test over discordant rules against \texttt{pymupdf4llm}.}
\label{tab:pdf}
\small
\begin{tabular}{lrrrrr}
\toprule
\textbf{cue} & $n$ & \textbf{ours} & \texttt{pymupdf4llm} & \textbf{plain} & $p$ \\
\midrule
\textbf{highlight} & 18 & \textbf{11}\,/\,11 & 4\,/\,12 & 0\,/\,15 & \textbf{0.039} \\
bold & 34 & 23\,/\,23 & 23\,/\,23 & 0\,/\,31 & 1.00 \\
italic & 47 & 14\,/\,22 & 19\,/\,25 & 0\,/\,40 & 0.125 \\
strikethrough & 27 & 17\,/\,19 & 19\,/\,22 & 0\,/\,21 & 0.500 \\
\midrule
\textbf{total} & 126 & \textbf{65}\,/\,75 & \textbf{65}\,/\,82 & 0\,/\,107 & 1.00 \\
\bottomrule
\end{tabular}
\end{table}

\section{Cross-Format Results: PDF}
\label{app:pdf}
\label{sec:eval_pdf}

If colour is the load-bearing cue, that should hold in a format encoded entirely
differently. PDF has no run properties: a highlight is a filled rectangle drawn behind
glyphs and a strikethrough is a hairline crossing the glyph midline, so cues must be
recovered geometrically from character bounding boxes. We evaluate on all 126 born-digital
formatting rules of ParseBench (Table~\ref{tab:pdf}).

Two things follow. First, plain PDF text extraction recovers \emph{no} cue at all
($0/126$) while locating $107$ of the $126$ targets---the same pattern as in Word, where
the information is present on the page and absent from the tokens. Second, we match the
strongest PDF baseline exactly on cue recovery ($65/126$ each, 9 discordant rules in each
direction), and the \emph{only} statistically distinguishable per-cue difference in either
direction is highlighting, where we lead $11/18$ against $4/18$ ($p=0.039$); bold is
identical, and our lower italic and strikethrough counts are not significant ($p=0.125$,
$p=0.500$). The cross-format picture therefore reproduces the Word result---broad parity,
with the colour-class cue as the differentiator---from a different encoding and scored by a
benchmark we did not write.

Two limitations. Our text \emph{localisation} trails \texttt{pymupdf4llm} ($75$ versus $82$
of $126$), so part of the italic and strikethrough shortfall is a reading-order deficit
rather than cue detection, and improving reading order would likely close it. And 18 of the
126 rules name text that is absent from the PDF's native text layer entirely: there,
\emph{every} arm recovers zero cues, and \texttt{pymupdf4llm} locates text on 4 of them
only by silently invoking OCR. We report those rules rather than excluding them, since
excluding them would change no cue number. Cue recovery for scanned pages, where the cue
survives as pixels but no text layer exists, needs OCR we do not perform and is left to
future work.

\section{Equal-Preprocessing Comparison}
\label{app:equalprep}
\label{sec:eval_system}

Every experiment so far hands a parsed representation straight to the answering model, which
invites the objection that we compared a full pipeline against bare converters. We therefore
ran the complete pipeline---an identical extraction prompt, model and tier logic, identical image
bytes for the vision pass, identical renderer---over 68 documents, varying
\emph{only} the Stage-1 text, and asked the same question of the rendered knowledge package
(Table~\ref{tab:equalprep}).

The advantage survives, and grows: $+0.178$ over docx2python (27/5/36, CI $[+0.080,+0.277]$)
and $+0.422$ over docling (40/2/26), both significant. So the effect is attributable to the
representation rather than to pipeline depth.

The retention column explains \emph{why}, and is the more informative result. All three arms
carry a similar share of the gold requirement text through extraction and rendering into the
final package---$0.665$, $0.634$, $0.607$---so the gap is not a matter of content being lost.
It is that only our package still records \emph{which} of those requirements the author
singled out. Salience survives the pipeline as placement: a marked requirement reaches the
quick-reference file at rate $0.569$ against $0.245$ for an unmarked one, a $2.3\times$
enrichment.

Absolute scores are lower than in \S\ref{sec:eval_main} for every arm, because extraction and
rendering deliberately compress and restructure, and because a $0.665$ retention rate caps
recall. We report the comparison rather than the absolute level. On faithfulness, a validator
sample of 24 entities on a held-out enterprise SOP found every entity traceable to a verbatim
source quote at $f(e)\!\ge\!2$ ($87.5\%$ fully faithful, the remainder under-specifying rather
than fabricating); Appendix~\ref{app:e2e} traces a complete run stage by stage.

\section{Cross-Model Generalisation}
\label{app:crossmodel}

\begin{table}[h]
  \centering
\caption{Cross-model generalisation, 76--79 documents per model. $\Delta$ columns are paired per-document macro F1 against the typed representation, exact sign test, Holm-corrected within each model. Plain text collapses under every model; beating docx2python replicates on four of six; the typing benefit replicates on three of six and is positive on five, never significantly negative. Claude Opus~4.6 values are from the six-arm run of Table~\ref{tab:flagship}.}
\label{app:tab:crossmodel}
\footnotesize
\setlength{\tabcolsep}{3pt}
\begin{tabular}{lrrrr@{\hskip 6pt}rr}
\toprule
& \multicolumn{4}{c}{\textbf{micro F1}} & \multicolumn{2}{c}{\textbf{$\Delta$ vs.\ ours} ($p_{\text{Holm}}$)} \\
\cmidrule(lr){2-5}\cmidrule(lr){6-7}
\textbf{Answering model} & ours & untyped & dx2py & plain & untyped & dx2py \\
\midrule
Claude Opus 4.6 & \textbf{0.687} & 0.605 & 0.586 & 0.075 & $+.097^{*}$ & $+.115^{*}$ \\
Claude Haiku 4.5 & 0.629 & 0.531 & 0.499 & 0.048 & $+.108^{**}$ & $+.168^{***}$ \\
Claude Sonnet 4.6 & 0.615 & 0.577 & 0.544 & 0.007 & $+.085^{*}$ & $+.122^{**}$ \\
GPT-5.6-terra & 0.534 & 0.468 & 0.510 & 0.045 & $+.019$ & $+.014$ \\
Claude Opus 5 & 0.358 & 0.346 & 0.263 & 0.068 & $+.015$ & $+.064^{*}$ \\
Claude Sonnet 5 & 0.164 & 0.173 & 0.180 & 0.018 & $+.013$ & $+.057$ \\
\bottomrule
\end{tabular}

\vspace{2pt}
{\scriptsize $^{*}p<0.05$, $^{**}p<0.01$, $^{***}p<0.001$.}
\end{table}

\section{Prevalence of Cue-Dependent Passages}
\label{app:prevalence}

\begin{table}[t]
  \centering
\caption{Prevalence of cue-dependent passages in 6{,}462 real documents. Detection is at OOXML run level. Pattern~D was invisible to earlier scans: a tracked deletion is a \texttt{w:r} nested inside \texttt{w:del} carrying \texttt{w:delText}, which neither \texttt{paragraph.runs} nor \texttt{iter\_inner\_content} reaches.}
\label{app:prevalence}
\footnotesize
\setlength{\tabcolsep}{4pt}
\begin{tabular}{llrr}
\toprule
& \textbf{Pattern} & \textbf{Passages} & \textbf{Documents} \\
\midrule
A & superseded value (struck + live) & 71 & 8 \\
B & struck content & 360 & 35 \\
C & colour/highlight on a constraint & 2{,}860 & 359 \\
D & tracked replacement & 68 & 21 \\
D & tracked deletion & 423 & 31 \\
\midrule
\multicolumn{2}{l}{\textbf{any pattern}} & --- & \textbf{414 (6.4\%)} \\
\bottomrule
\end{tabular}
\end{table}

\section{Architectural Design Principles}
\label{app:principles}

Five principles guide the system's architecture:

\begin{enumerate}[leftmargin=*,nosep]
    \item \textbf{LLM for reasoning, code for assembly.} LLM calls handle tasks requiring judgment (entity classification, relationship inference, editorial summarization). Deterministic tasks (parsing, serialization, rendering, file I/O) remain in code. This minimizes cost, maximizes reproducibility, and reduces non-determinism.

    \item \textbf{Formatting is signal, not decoration.} Every formatting attribute (bold, italic, color, highlight, font) is preserved through the pipeline and presented to the LLM as a semantic hint. The pipeline never strips formatting.

    \item \textbf{Source-quote provenance.} Every extracted entity must cite a verbatim quote from the source document. This is the primary anti-hallucination mechanism, more effective than any post-hoc validation.

    \item \textbf{Type specificity over generality.} Seventeen entity types with distinct schemas produce richer extractions than a generic ``extract all entities'' approach. A \textsc{Procedure} has steps with actors and deadlines; a \textsc{Rule} has conditions and consequences. These distinctions matter for downstream use.

    \item \textbf{Progressive disclosure over monolithic output.} The output is not a single flat document but a tiered package: a lean entry point containing critical guardrails, a quick-reference sheet for common lookups, and typed sub-files loaded on demand. This matches how AI assistants actually consume skill content.
\end{enumerate}

\section{Complete Formatting-to-LLM Pipeline}
\label{app:pipeline}
\label{app:formatting_pipeline}

Table~\ref{tab:formatting_pipeline} details the complete formatting-to-LLM pipeline for all 11 active SemanticRoles introduced in Table~\ref{tab:semantic_roles}.

\begin{table}[t]
  \centering
\caption{Complete formatting-to-LLM pipeline for all 11 HIGH-confidence SemanticRoles. \textbf{Disambiguation Rule} is the parse-time algorithm; \textbf{Serialization} is what the LLM receives verbatim; \textbf{Extraction Signal} is the semantic intent conveyed. Block-level roles wrap the entire paragraph via \texttt{[PARAGRAPH roles=\ldots]} attributes; span-level roles wrap individual runs inline. Medium-confidence rules (italic $\to$ document reference, green text $\to$ status marker) are excluded: a wrong hint biases the LLM toward an incorrect type.}
\label{tab:formatting_pipeline}
\small
\setlength{\tabcolsep}{4pt}
\begin{tabular}{p{1.6cm}p{3.8cm}p{2.5cm}p{3.2cm}p{3.6cm}}
\toprule
\textbf{Word Attr.} & \textbf{Disambiguation Rule} & \textbf{SemanticRole} & \textbf{LLM Serialization} & \textbf{Extraction Signal} \\
\midrule
Bold span &
  Some but not all spans in the block are bold &
  \textsc{Key\_Term} &
  \texttt{**text**} (Markdown) &
  Definitional term or key concept; strong candidate for entity name \\
\addlinespace
Bold paragraph &
  Block is standalone (single line, $<$200 chars) and every span is bold &
  \textsc{Implicit\_Heading} &
  \texttt{[PARAGRAPH roles=} \texttt{"IMPLICIT\_HEADING"]} &
  Section boundary; treat following content as a sub-section even without an explicit heading style \\
\addlinespace
ALL-CAPS line &
  Every character run in the paragraph is all-caps; no mixed-case text present &
  \textsc{Implicit\_Heading} &
  \texttt{[PARAGRAPH roles=} \texttt{"IMPLICIT\_HEADING"]} &
  Implicit section label used by some authors instead of styled headings; same signal as above \\
\addlinespace
Italic / underline &
  No disambiguation needed; single HIGH-confidence mapping &
  \textsc{Emphasis} &
  \texttt{*text*} (Markdown) &
  Emphasized qualifier; often an exception trigger, a deadline phrase, or a key condition \\
\addlinespace
Red text &
  Fuzzy threshold: $\mathit{is\_red}(r,g,b) = (r{>}180) \wedge (g{<}80) \wedge (b{<}80)$, plus a set of known red hex literals (\#FF0000, \#C00000, \#CC0000). Tolerates RGB variance from custom author palettes &
  \textsc{Critical\_Warning} &
  \texttt{[COLOR:red]text[/COLOR]} (custom tag, block attr.\ if whole paragraph) &
  Critical prohibition, safety constraint, or immediate-consequence warning; always extract as \textsc{Warning} with \texttt{CRITICAL} severity \\
\addlinespace
Yellow highlight &
  Highlight attribute present and not \texttt{WHITE}/\texttt{NONE}; \texttt{RED} escalates to \textsc{Critical\_Warning} &
  \textsc{Warning} &
  \texttt{[HIGHLIGHT:}\textit{hue}\texttt{]} \texttt{text[/HIGHLIGHT]} &
  Caution statement; extract as \textsc{Warning} with \texttt{MAJOR} severity if an explicit consequence is stated \\
\addlinespace
Strikethrough &
  Exact formatting attribute; no disambiguation &
  \textsc{Deprecated} &
  \texttt{\textasciitilde\textasciitilde text\textasciitilde\textasciitilde} (Markdown) &
  Superseded policy value; do not extract as current constraint; flag for conflict detection \\
\addlinespace
Monospace font &
  Font family name in known monospace set (Courier, Consolas, Lucida Console, Menlo, Monaco \ldots) &
  \textsc{Code\_or\_Literal} &
  \texttt{`code`} (Markdown inline code) &
  Exact system identifier, form number, command, or literal value; preserve verbatim \\
\addlinespace
Boxed / bordered &
  Paragraph border attribute present on all four sides, or background shading applied &
  \textsc{Callout} &
  \texttt{[PARAGRAPH roles="CALLOUT"]} &
  Reference callout or tip box; supporting information, not an operative guardrail \\
\addlinespace
Superscript run &
  Superscript text property set on a character run &
  \textsc{Reference\_Marker} &
  \texttt{[PARAGRAPH roles=} \texttt{"REFERENCE\_MARKER"]} &
  Footnote or citation marker; track for relationship extraction, not a standalone entity \\
\addlinespace
Header + pattern &
  Header/footer text matches \texttt{/DRAFT|CONFIDENTIAL|v\textbackslash d+\textbackslash ./} (regex on document headers) &
  \textsc{Metadata} &
  \texttt{[PARAGRAPH roles="METADATA"]} &
  Document lifecycle metadata (version, classification); exclude from entity extraction \\
\bottomrule
\end{tabular}
\end{table}

\section{Serialization Tag Set and Attribute Encoding}
\label{app:serialization_details}

\textbf{Structural metadata tags.} Every block is wrapped in a structural tag carrying document-level metadata:
\begin{itemize}[leftmargin=*,nosep]
    \item \texttt{[HEADING level=2 style="Heading 2"]} --- heading level and Word style name
    \item \texttt{[TABLE has\_header=True cols=4]} --- table structure hints
    \item \texttt{[IMAGE type="flowchart" size=45K]} --- image classification placeholder
    \item \texttt{[PARAGRAPH roles="CRITICAL\_WARNING"]} --- block-level roles
\end{itemize}
These structural tags encode information the LLM cannot infer from text alone: which numbered paragraphs are part of an ordered list, which table has a header row, which paragraph sits under a given heading level. Without these, the extraction model would need to re-derive heading hierarchy from font size heuristics, a fragile and error-prone operation.

\textbf{Attribute encoding for fine-grained context.} Tag attributes also carry indentation depth (for nested lists), font size (to distinguish normal body text from large callout headings), and style name (which encodes the Word template's intended semantic, e.g., ``Caption'', ``List Number 2''). Style names are particularly valuable because they reflect the document author's own semantic intent, which often aligns with entity type triggers.

\section{Running Example: Parsing an Onboarding SOP}
\label{app:parsing_example}

To illustrate the pipeline concretely, we use a running example throughout: an ``Employee Onboarding at Acme Corp'' SOP document. Consider the following Word content:

\begin{lstlisting}[style=textstyle,caption={Raw Word content with formatting (bold and red indicated).}]
[Heading 1, bold, 16pt]
1. Introduction

[Normal paragraph]
A <<bold>>probationary period<</bold>> is the first
90 days of employment during which the employee's
performance and fit are evaluated. During this
period, either party may terminate employment with
5 days' written notice.

[Normal paragraph, red text, bold]
WARNING: Employees who fail to submit the I-9 form
within 3 business days will be immediately suspended
without pay.
\end{lstlisting}

After Stage~1 parsing and serialization, this becomes:

\begin{lstlisting}[style=textstyle,caption={Annotated text produced by Stage 1. Bold uses Markdown; red uses custom tag; block-level role appears as tag attribute.}]
[HEADING level=1 style="Heading 1"]
1. Introduction
[/HEADING]

[PARAGRAPH style="Normal"]
A **probationary period** is the first 90 days of
employment during which the employee's performance
and fit are evaluated. During this period, either
party may terminate employment with 5 days' written
notice.
[/PARAGRAPH]

[PARAGRAPH style="Normal" roles="CRITICAL_WARNING"]
[COLOR:#FF0000]**WARNING: Employees who fail to
submit the I-9 form within 3 business days will be
immediately suspended without pay.**[/COLOR]
[/PARAGRAPH]
\end{lstlisting}

The parser has: (a) preserved bold as Markdown \texttt{**...**}, (b) serialized red text as \texttt{[COLOR:\#FF0000]...[/COLOR]}, (c) assigned the \textsc{Critical\_Warning} semantic role to the red paragraph at the block level (the entire paragraph is red, not just one span), and (d) maintained structural metadata (heading level, style name).

\section{Handling python-docx Limitations}
\label{app:docx_limits}

Several content types common in enterprise documents are not natively exposed by \texttt{python-docx} and require XML-level access:

\begin{itemize}[leftmargin=*,nosep]
    \item \textbf{Text boxes}: Content in \texttt{w:txbxContent} elements must be extracted by walking the XML tree directly.
    \item \textbf{Tracked changes}: \texttt{python-docx} silently skips deleted text. Insertions (\texttt{w:ins}) and deletions (\texttt{w:del}) require XML parsing.
    \item \textbf{Comments}: Stored in \texttt{word/comments.xml}, accessible only through the OPC package interface.
    \item \textbf{Merged cells}: Vertical merge continuations (\texttt{tc.vMerge}) report as empty cells.
    \item \textbf{OLE objects}: Embedded objects appear as empty paragraphs; detected via \texttt{w:object} elements and logged as warnings.
\end{itemize}

These workarounds add approximately 80 lines of XML-handling code but are essential for complete extraction from real-world enterprise documents.

\section{Why 17 Entity Types?}
\label{app:whyseventeen}
\label{app:why17}

Fewer types (e.g., the 6-type schema common in generic IE) force-fit distinct content into overly broad categories: a \textsc{Rule} with conditions and consequences is fundamentally different from a \textsc{Procedure} with ordered steps, even though both describe ``what to do.'' The 17-type taxonomy was validated against real enterprise SOPs to ensure each type has clear, non-overlapping trigger signals.

\section{The 12-Point Extraction Mandate}
\label{app:mandate}

Beyond the obvious ``extract all entities'' instruction, several rules address specific failure modes observed in early testing:

\begin{itemize}[leftmargin=*,nosep]
    \item \textit{Read-before-extract}: The model must read the entire document before beginning extraction, preventing the common failure mode where entities in later sections are missed because the model starts outputting before fully reading.
    \item \textit{Checklist-as-procedure}: Checkbox lists are \textsc{Procedure} entities, not \textsc{Warning} entities, even if they contain prohibitions. Without this rule, models frequently misclassify ``checklist items to verify'' as warnings.
    \item \textit{WRONG/CORRECT pattern}: When a document presents ``Incorrect: X / Correct: Y'' examples, these become \texttt{worked\_examples} in the \texttt{type\_specific\_fields} of a \textsc{Rule} entity. Without this rule, the WRONG examples are sometimes extracted as independent factual claims.
    \item \textit{Procedural completeness}: Sub-steps within a step must be preserved in full. The model must never ``summarize'' a 12-step sub-procedure into a one-line description. This rule prevents the single most common form of information loss in SOP extraction.
    \item \textit{Warnings and prohibitions are highest priority}: Even if the surrounding content is low-confidence, a \textsc{Warning} entity with red text and explicit consequences must always be extracted at HIGH confidence.
\end{itemize}

The remaining 7 points of the 12-point mandate cover standard but necessary rules: ALL-CAPS lines as implicit headings, proper \textsc{Enumeration} item extraction (preserve all items verbatim), \textsc{Scope} entity completeness, \textsc{Metric} unit and threshold preservation, cross-reference tracking for relationship extraction, confidence calibration for diagram-derived entities, and deduplication precedence (explicit stated entity wins over diagram-inferred).

\section{Few-Shot Example Selection}
\label{app:fewshot}

The five few-shot examples included in the extraction prompt, each targeting a specific failure mode:

\begin{enumerate}[leftmargin=*,nosep]
    \item A \textsc{Definition} with bold formatting (demonstrates key term detection from inline bold).
    \item A \textsc{Procedure} from a numbered list (demonstrates step extraction with actors and deadlines).
    \item A complex paragraph yielding \textsc{Rule} + \textsc{Exception} + \textsc{Form\_or\_Template} simultaneously (multi-entity extraction from a single passage---the most common source of extraction errors).
    \item A \textsc{Procedure} with nested sub-steps and an embedded \textsc{Warning} (demonstrates that sub-steps must be preserved in full and warnings extracted as siblings, not merged into the procedure).
    \item A \textsc{Rule} with WRONG/CORRECT worked examples (demonstrates that comparison tables encode the rule's \texttt{worked\_examples} field, not separate factual entities).
\end{enumerate}

\section{Running Example: Extraction Output}
\label{app:extraction_example}

Continuing our onboarding SOP example, the bold term ``\textbf{probationary period}'' triggers \textsc{Definition} extraction:

\begin{lstlisting}[style=jsonstyle,caption={Extraction output for a \textsc{Definition} entity. Source quote enables faithfulness verification.}]
{
  "id": "e001",
  "type": "DEFINITION",
  "name": "Probationary Period",
  "description": "A 90-day initial employment period
    for evaluating new hire performance and fit,
    with shortened termination notice (5 days vs.
    standard 30 days).",
  "summary": "90-day new hire evaluation period.",
  "source_quotes": ["A **probationary period** is the
    first 90 days of employment..."],
  "confidence": "HIGH",
  "type_specific_fields": {
    "definition": "The first 90 days of employment
      during which performance and fit are evaluated.",
    "key_facts": ["Duration: 90 days",
      "5-day notice during probation"],
    "domain": "HR"
  }
}
\end{lstlisting}

The red-text warning triggers a \textsc{Warning} extraction with CRITICAL severity driven by both the \textsc{Critical\_Warning} semantic role (from the red text annotation) and the explicit consequence (immediate suspension):

\begin{lstlisting}[style=jsonstyle,caption={\textsc{Warning} extraction driven by red text semantic role and consequence severity.}]
{
  "id": "e025",
  "type": "WARNING",
  "name": "I-9 Submission Deadline Violation",
  "confidence": "HIGH",
  "type_specific_fields": {
    "severity": "CRITICAL",
    "message": "I-9 form must be submitted within
      3 business days of start date.",
    "consequence_of_ignoring": "Immediate suspension
      without pay"
  }
}
\end{lstlisting}

A more complex passage demonstrates multi-entity extraction and the WRONG/CORRECT pattern. The passage on compliance training yields a \textsc{Rule}, \textsc{Exception}, and \textsc{Form\_or\_Template} from a single paragraph---along with two typed relationships (\textsc{Exempts} and \textsc{References}) extracted in the same pass.

\section{Structured Output via Tool Use}
\label{app:tool_use}

\textbf{Free-form type\_specific\_fields.} The tool use schema intentionally leaves \texttt{type\_specific\_fields} as a free-form JSON object rather than defining a per-type nested schema. This avoids an excessively deep and brittle tool definition ($>$200 lines of JSON Schema) while still guiding the model through the system prompt's per-type field descriptions. The Pydantic model that receives the output validates only the required fields, tolerating per-type variation.

The system uses the Strands Agents SDK with a \texttt{@strands.tool}-decorated function as the extraction target. The model is forced to call \texttt{extract\_entities} with the full \texttt{ExtractionResult} payload, which Pydantic validates before returning to the pipeline. This guarantees valid structured output without JSON parsing heuristics. The agent's message history is inspected post-call to extract the tool input dict, which is then deserialized into typed Python models.

\section{Tier 3 Merge: Additional Details}
\label{app:merge}
\label{app:merge_details}

\textbf{Segmentation walk.} The algorithm walks blocks sequentially, accumulating a running token count and starting a new segment when the 8,000-token threshold is exceeded.

\textbf{Oversized block handling during segmentation.} A subtle complication arises with oversized blocks---single paragraphs exceeding $2\times$ the overlap budget (i.e., $>$8,000 characters). Including these in the overlap would cause them to dominate the next segment's context, crowding out the actual boundary content. Such blocks are explicitly excluded from overlap computation via a \texttt{skip\_next\_overlap} flag, ensuring the overlap consists of the intended preceding context rather than one large anomalous block.

\textbf{Restore from originals.} After merging, the compact summaries are replaced with full original data by matching merged entity names to original segment entities via case-insensitive string matching. For entities that appeared in multiple segments (cross-segment duplicates), the merge call identifies them as a single entity; the restore step takes the union of all \texttt{source\_quotes} arrays and merges \texttt{type\_specific\_fields} (e.g., unioning steps arrays from both segments ensures no sub-steps are lost during deduplication).

\section{Why Source-Quote Verification Instead of Mutual Dependency?}
\label{app:md}

The ZOES Mutual Dependency (MD) principle validates (entity, attribute, value) triplets by checking if masking one component allows recovery of the others. MD addresses \emph{attribute granularity}---ensuring discovered attributes are neither too coarse nor too fine. In our guided-schema extraction, attributes are predefined, so the granularity problem MD solves does not exist. The actual risk is \emph{hallucination}: the LLM fabricating content not in the source. Source-quote verification targets this risk directly with a simpler prompt and lower cost.

\section{Diagram Taxonomy and Supporting Protocol Figures}
\label{app:figures}

\begin{figure}[!htbp]
    \centering
    \includegraphics[width=0.75\textwidth]{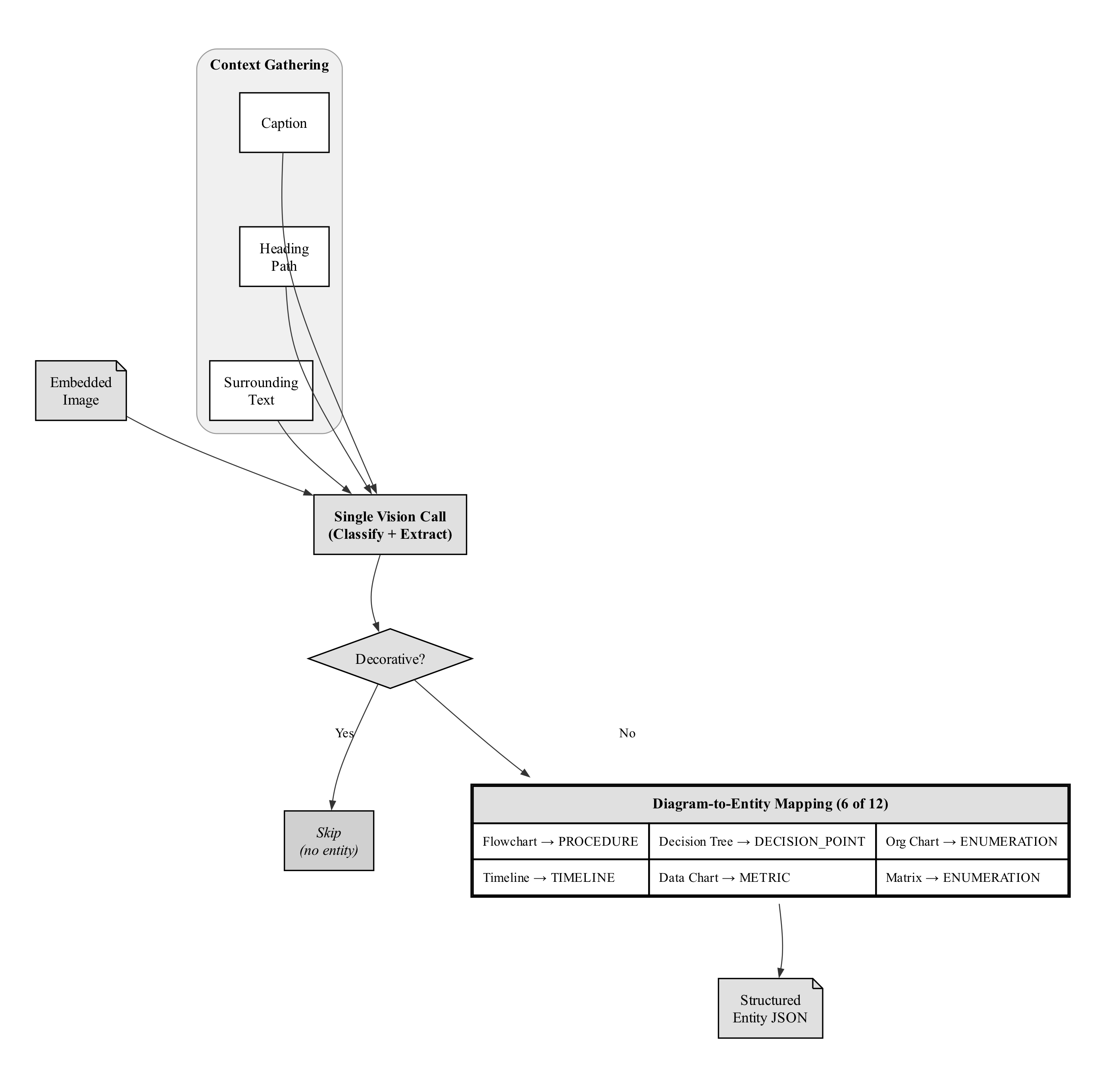}
    \caption{Single-pass diagram understanding protocol. Context from surrounding text (caption, heading path, adjacent paragraphs) is provided alongside the image in a single vision call. The model classifies the diagram type and extracts entities simultaneously. Vision calls run sequentially after text extraction, with decorative images skipped early.}
    \label{fig:diagram_protocol}
\end{figure}

\begin{figure}[!htbp]
    \centering
    \includegraphics[width=\columnwidth]{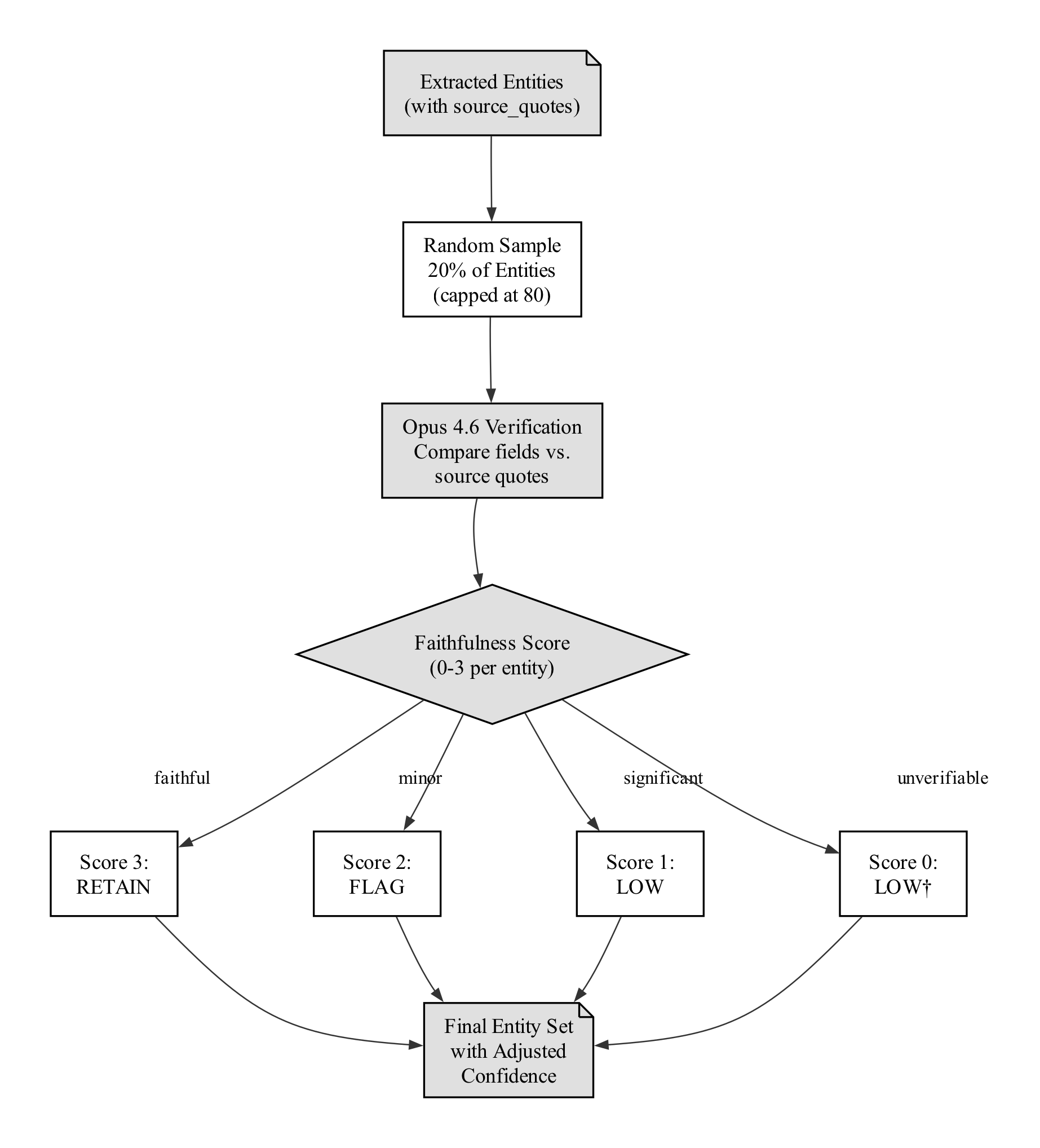}
    \caption{Faithfulness verification flow. A 20\% random sample (capped at 80 entities) is verified against source quotes using Opus 4.6. Score 0 or 1 downgrades confidence to \textsc{Low}; score 0 does not discard (to avoid false-positive losses from truncation). Relationships referencing downgraded entities are also updated.}
    \label{fig:faithfulness_check}
\end{figure}

Table~\ref{tab:diagram_types} lists the 12 diagram categories and their entity mappings. Figure~\ref{fig:diagram_protocol} illustrates the single-pass diagram understanding protocol with context gathering from surrounding text. Figure~\ref{fig:faithfulness_check} illustrates the faithfulness verification flow.

\begin{table}[h]
  \centering
\caption{Diagram type categories with entity mappings.}
\label{tab:diagram_types}
\small
\begin{tabular}{lll}
\toprule
\textbf{Diagram Type} & \textbf{Primary Entity} & \textbf{Secondary} \\
\midrule
Flowchart & \textsc{Procedure} & \textsc{Rule}, \textsc{Role} \\
Decision Tree & \textsc{Decision\_Point} & \textsc{Rule} \\
Org Chart & \textsc{Enumeration} & \textsc{Role} \\
Matrix (RACI) & \textsc{Enumeration} & \textsc{Role} \\
Timeline & \textsc{Timeline} & \textsc{Rule} \\
Data Chart & \textsc{Metric} & -- \\
Architecture & \textsc{Definition} & \textsc{Reference} \\
Conceptual & \textsc{Definition} & \textsc{Reference} \\
Table as Image & \textsc{Enumeration} & -- \\
Form & \textsc{Form\_or\_Template} & \textsc{Procedure} \\
Screenshot & \textsc{Scope} & -- \\
Decorative & \textit{(skipped)} & -- \\
\bottomrule
\end{tabular}
\end{table}

\section{Progressive Disclosure: Package Layout and File Roles}
\label{app:disclosure}

\begin{lstlisting}[style=textstyle,caption={Multi-file output package structure. Each file serves a distinct cognitive role.}]
SKILL.md                      # Entry point: <500 lines
QUICK_REFERENCE.md            # Auto-generated cheat sheet
docs/
  schema-reference.md         # Large ENUMERATIONs
  concepts-and-metrics.md     # DEFINITIONs, METRICs, GLOSSARYs
  procedures-reference.md     # Long PROCEDUREs, error-handling
references/
  provenance.json             # Entity ID -> source quote map
\end{lstlisting}

\textbf{SKILL.md (entry point).} Contains the YAML frontmatter, a ``How to Use This Skill'' section linking to sub-files, and only the action-critical content: scope, prerequisites, exceptions, bare guardrail rules, warnings, and short procedures/enumerations. Strict size target: $<$500 lines. Entities routed to sub-files appear as one-line stubs with a ``See \texttt{docs/X.md}'' link.

\textbf{QUICK\_REFERENCE.md (auto-generated cheat sheet).} Auto-generated from all \textsc{Warning} entities and high-severity \textsc{Rule} entities. Contains severity annotations, consequence statements, and mitigation guidance. Only generated when such entities exist. This file is the first reference an AI assistant should consult---it surfaces the most operationally critical constraints without requiring navigation into deeper sub-files.

\textbf{docs/ sub-files (full reference content).} Each sub-file uses the same hierarchy-aware rendering as SKILL.md but with full entity content: complete procedure steps with actor annotations, full enumeration tables with attribute columns, metric formulas and targets, glossary terms and definitions. Sub-files are loaded on demand, keeping the main context window lean.

\textbf{references/provenance.json.} Maps every entity ID to its source quotes, source section path, and extraction confidence. Enables complete auditability: any AI-generated statement derived from the skill can be traced to a verbatim source passage.

\section{Anthropic-Spec Description Generation}
\label{app:description}

The SKILL.md YAML frontmatter requires a \texttt{description} field that the Anthropic skill specification calls ``pushy''---it must use imperative trigger language to combat undertriggering (the failure mode where the AI assistant possesses the skill but fails to load it because the user's query doesn't explicitly mention the skill name).

The description generation algorithm uses a three-tier fallback:

\begin{enumerate}[leftmargin=*,nosep]
    \item \textit{Audience + entity trigger keywords (primary)}: Extract \texttt{applies\_to} from the \textsc{Scope} entity as audience context. Collect up to 5 \textsc{Procedure}, \textsc{Rule}, and \textsc{Warning} entity names as trigger keywords. Format: ``For \{audience\}. Use this skill whenever users ask about \{entity\_names\}.''\footnote{Entity names from the document are preferred over \texttt{applies\_to} as triggers because they match the vocabulary users employ when asking questions (``How do I reset a password?'') rather than the document's audience description (``IT Support team handling end-user requests'').}
    \item \textit{SCOPE entity summary (fallback)}: If no \textsc{Scope} entity has \texttt{applies\_to}, use the scope entity's LLM-generated summary as the description.
    \item \textit{Domain + high-confidence entity names (last resort)}: Use the document title plus names of HIGH-confidence \textsc{Definition}, \textsc{Rule}, and \textsc{Procedure} entities.
\end{enumerate}

\section{Running Example: Rendered Output}
\label{app:rendering_example}

The extracted entities from our onboarding SOP produce the following SKILL.md structure:

\begin{lstlisting}[style=textstyle,caption={Rendered SKILL.md excerpt. Frontmatter has name and description only (Anthropic spec). No type labels in entity headers. Long procedure appears as stub with sub-file link.}]
---
name: acme-employee-onboarding
description: For all new hires and HR personnel.
  Use this skill whenever users ask about I-9
  Verification Process, Compliance Training
  Deadline, Benefits Enrollment Procedure.
---

## How to Use This Skill

Covers the complete onboarding workflow for
new hires joining Acme Corp.

When answering questions:
- Consult QUICK_REFERENCE.md first -- critical
  warnings and compliance deadlines
- See docs/procedures-reference.md for step-by-step
  procedures with actor annotations
- See docs/concepts-and-metrics.md for definitions,
  role descriptions, and timeline milestones

## Overview and Scope

**Acme Corp Employee Onboarding**

Covers all new full-time and part-time employees
hired on or after January 1, 2024.

**Applies to**: All new hires, HR Business Partners,
  Direct Managers

## Day One Procedures

**I-9 Verification Process**

Federal I-9 verification requiring Section 1
completion by the employee on or before Day 1,
and Section 2 completion by HR within 3 days.
-> See docs/procedures-reference.md

> **CRITICAL**: I-9 form must be submitted within
> 3 business days. Non-compliance triggers immediate
> suspension without pay.
\end{lstlisting}

Note the key rendering decisions: (a) \texttt{name} and \texttt{description} are the only frontmatter fields (Anthropic spec); (b) entity headers carry no type labels (removed as presentation noise); (c) the long \textsc{Procedure} appears as a one-line stub with a sub-file link; (d) the \textsc{Warning} is rendered as a blockquote with CRITICAL severity inline in \texttt{SKILL.md}.

\section{Open-Schema Discovery}
\label{app:openschema}

The 17-type taxonomy covers approximately 95\% of enterprise SOP content. For novel document types containing structures outside the taxonomy (e.g., escalation matrices, audit checklists), the system supports an optional open-schema discovery mode inspired by ZOES's Root Attribute Induction~\cite{xu2026zoes}.

The approach uses a two-pass prompt strategy: Pass~1 adds the instruction ``if you encounter structured information that does not fit any of the 17 types, create a new type with a descriptive name and define its schema'' to the standard extraction prompt. If novel types are discovered, Pass~2 re-extracts with the expanded taxonomy to ensure consistent application across the entire document.

\section{Conflict Resolution}
\label{app:conflict}

When the source document contains contradictory information (e.g., ``probationary period is 90 days'' in Section~1 vs. ``60 days for part-time employees'' in Section~7), the system: (1) flags the entity with a conflict marker, (2) preserves \emph{both} values with source citations, (3) adds an entry to a dedicated ``Conflicts \& Ambiguities'' section in the output, and (4) lowers the entity's confidence. The system never silently resolves contradictions---that decision belongs to a human reviewer.

\section{Implementation Details}
\label{app:implementation}

\textbf{Architecture and dependencies.} The system is implemented as a Python 3.12+ package comprising $\sim$1,800 lines of production code across 10 modules. The largest modules are: \texttt{llm/prompts.py} ($\sim$763 lines, extraction + merge + validation prompts), \texttt{rendering/skill\_renderer.py} ($\sim$1,100 lines, entity routing and multi-file rendering), and \texttt{llm/extraction.py} ($\sim$630 lines, tier orchestration, segmentation, and binary merge).

The five runtime dependencies are: \texttt{python-docx} ($\geq$0.8.11) for document parsing, an agent SDK ($\geq$0.1.0) for LLM calls against a managed cloud LLM service, \texttt{pydantic} ($\geq$2.0) for structured output validation, \texttt{click} ($\geq$8.0) for CLI, and \texttt{rich} ($\geq$13.0) for progress display. Notably absent: NumPy, sentence-transformers, FAISS, scikit-learn---the embedding-based deduplication machinery common in document processing pipelines is entirely replaced by inline LLM deduplication.

\textbf{Bedrock integration.} LLM calls use the Strands Agents SDK's \texttt{BedrockModel} with cross-region inference profiles (\texttt{us.anthropic.claude-opus-4-6-v1} for all calls: extraction, merge, faithfulness verification, and vision). The SDK handles streaming, retry with exponential backoff ($\text{delay} = 2^{\text{attempt}} + \text{jitter}$), and structured output via tool use.

Errors are classified as retryable (throttling, rate limits, transient server errors, connection timeouts) or non-retryable (authentication failures, expired tokens, malformed requests, maximum token exceeded). On retry exhaustion after 3 attempts, a \texttt{PipelineError} is raised with context about the failed tier and call index, enabling targeted debugging.

\section{End-to-End Example}
\label{app:e2e}

This appendix traces the complete pipeline execution on the included test document, \texttt{test\_doc.docx}---an IT Password Reset Procedure SOP. All inputs and outputs shown are real data from an actual pipeline run.

\subsection{Stage 1 Input: Source Document Structure}

The input is a Word document with 8 sections: Scope, Prerequisites, Password Reset Steps, Security Warning, Account Lockout Policy, Compliance Logging Rule, Service Level Targets, and Glossary. The document exercises three key formatting signals:

\begin{itemize}[leftmargin=*,nosep]
  \item \textbf{Bold runs}: ``\texttt{password reset process}'', step labels (``\texttt{Log in:}'', ``\texttt{Locate account:}'', etc.), policy terms (``\texttt{Lockout threshold:}'', ``\texttt{1 hour}'').
  \item \textbf{Red text} (RGB 255, 0, 0): The entire Security Warning paragraph triggers \textsc{Critical\_Warning} role.
  \item \textbf{Heading styles}: Heading 1/2 triggers heading hierarchy reconstruction.
\end{itemize}

\subsection{Stage 1 Output: Hybrid Serialization (excerpt)}

The parser serializes the warning paragraph with its semantic role tag:

\begin{lstlisting}[style=textstyle]
[HEADING level=2]Security Warning[/HEADING]
[COLOR:red][CRITICAL_WARNING]NEVER share temporary passwords via email
or chat messages. Email is not a secure channel and violates the
company security policy. Always use a phone call or an encrypted
messaging platform.[/CRITICAL_WARNING][/COLOR]
\end{lstlisting}

Bold key terms in the lockout section are annotated inline:

\begin{lstlisting}[style=textstyle]
[HEADING level=1]Account Lockout Policy[/HEADING]
**Lockout threshold:** Accounts are automatically locked after
**5 consecutive failed login attempts**.
**Lockout duration:** 30 minutes. IT admins may unlock accounts
manually before the window expires.
\end{lstlisting}

\subsection{Stage 2: Tier Selection and Extraction}

Token estimation: $\sim$579 tokens $\Rightarrow$ \textbf{Tier~1} (single LLM call). The Opus 4.6 extraction call completes in $\sim$92 seconds and returns 20 entities and 16 relationships with 0 conflicts.

\textbf{Entity distribution:}

\begin{table}[h]
  \centering
\small
\begin{tabular}{lr|lr}
\toprule
\textbf{Type} & \textbf{Count} & \textbf{Type} & \textbf{Count} \\
\midrule
\textsc{Rule} & 5 & \textsc{Definition} & 1 \\
\textsc{Prerequisite} & 3 & \textsc{Metric} & 3 \\
\textsc{Procedure} & 1 & \textsc{Exception} & 1 \\
\textsc{Warning} & 1 & \textsc{Scope} & 1 \\
\textsc{Glossary\_Entry} & 3 & \textsc{Concept} & 1 \\
\bottomrule
\end{tabular}
\end{table}

\textbf{Example extracted entity} (entity e006, type \textsc{Warning}):

\begin{lstlisting}[style=jsonstyle]
{
  "id": "e006",
  "name": "Never Share Temporary Passwords via Email or Chat",
  "type": "WARNING",
  "confidence": "HIGH",
  "description": "Critical security prohibition against sharing
    temporary passwords via email or unsecured chat.",
  "severity": "CRITICAL",
  "warning_message": "NEVER share temporary passwords via email
    or chat messages.",
  "consequence": "Violation of company security policy;
    potential unauthorized access",
  "mitigation": "Always use a phone call or an encrypted
    messaging platform (e.g., secure Slack DM)",
  "source_sections": ["Security Warning"],
  "source_quotes": ["NEVER share temporary passwords via email
    or chat messages. Email is not a secure channel and
    violates the company security policy..."]
}
\end{lstlisting}

\textbf{Example extracted entity} (entity e005, type \textsc{Procedure}):

\begin{lstlisting}[style=jsonstyle]
{
  "id": "e005",
  "name": "Password Reset Steps",
  "type": "PROCEDURE",
  "confidence": "HIGH",
  "description": "Six-step AD password reset: log in, locate,
    reset, set complexity, force change, notify.",
  "steps": [
    {"step": 1, "actor": "IT Support technician",
     "action": "Log in to the Active Directory Management
       Console using your admin credentials."},
    {"step": 2, "action": "Navigate to Users and Computers,
       then search for the user account by name or email."},
    {"step": 3, "action": "Right-click the user account and
       select 'Reset Password'."},
    {"step": 4, "action": "Enter a temporary password that
       meets complexity requirements: minimum 12 chars..."},
    {"step": 5, "action": "Check 'User must change password
       at next logon' before clicking OK."},
    {"step": 6, "action": "Communicate the temporary password
       to the user by phone or secure Slack DM. Never email."}
  ],
  "source_quotes": ["Log in to the Active Directory...", ...]
}
\end{lstlisting}

\subsection{Stage 3 Output: Progressive Disclosure Package}

The content-analysis router distributes all 20 entities across 5 output files:

\begin{itemize}[leftmargin=*,nosep]
  \item \textbf{SKILL.md} (113 lines): Scope, all 3 prerequisites, 4 rules (bare constraints), 1 warning, 1 exception. The \textsc{Procedure} entity (6 steps) is routed to \texttt{docs/} and replaced with a stub link.
  \item \textbf{QUICK\_REFERENCE.md}: Auto-generated from the 1 \textsc{Warning} entity + 4 critical \textsc{Rule} entities, formatted for instant lookup.
  \item \textbf{docs/procedures-reference.md}: Full 6-step password reset procedure with actor, prerequisites, and postconditions.
  \item \textbf{docs/concepts-and-metrics.md}: 1 \textsc{Definition}, 3 \textsc{Metric} entities (resolution time, SLA, audit cycle), 3 \textsc{Glossary\_Entry} entities.
  \item \textbf{references/provenance.json}: Full source-quote map for all 20 entities.
\end{itemize}

\textbf{SKILL.md frontmatter} (auto-generated):

\begin{lstlisting}[style=textstyle]
---
name: it-password-reset-procedure
description: For All Active Directory accounts, IT Support team
  handling end-user requests. Use this skill whenever users ask
  about Password Reset Steps, Never Share Temporary Passwords
  via Email or Chat, Account Lockout Threshold Policy,
  Compliance Logging Rule for Password Resets, Password
  Complexity Requirements.
---
\end{lstlisting}

The description uses the SCOPE's \texttt{applies\_to} field as audience prefix and the top-priority entity names as trigger keywords---following the Anthropic Skill spec's ``pushy'' imperative style to maximize retrieval recall.

\textbf{QUICK\_REFERENCE.md excerpt} (auto-generated):

\begin{lstlisting}[style=textstyle]
## Critical Warnings

NEVER share temporary passwords via email or chat messages.
Email is not a secure channel and violates company policy.
Always use phone or encrypted messaging.

## Key Rules

- Account lockout: auto-locked for 30 min after 5 failed attempts
- Compliance: log all resets in ticketing system within 1 hour
- Password complexity: min 12 chars, upper+lower+digit+symbol
- Force change: check "must change at next logon" before OK
\end{lstlisting}

The 579-token source document produces a 5-file structured package. The SKILL.md entry point is 113 lines (under the 500-line target), with full provenance preserved in \texttt{references/provenance.json}.

\end{document}